\pdfoutput=1

\documentclass[11pt, dvipsnames]{article}

\usepackage[preprint]{acl}

\usepackage{times}
\usepackage{latexsym}

\usepackage[T1]{fontenc}

\usepackage[utf8]{inputenc}

\usepackage{microtype}

\usepackage{inconsolata}

\usepackage{graphicx}

\usepackage{hyperref}
\usepackage{tabularx,booktabs,caption}
\usepackage{setspace}
\usepackage{linguex}
\usepackage{natbib}
\usepackage{arydshln}
\usepackage{tikz-qtree}
\usepackage{amsmath}
\usepackage{amssymb}
\usepackage{bbding}
\usepackage{pifont}
\usepackage{booktabs}
\usepackage{lipsum}
\usepackage{csquotes}
\usepackage{comment}
\usepackage{endnotes}
\usepackage{xspace}
\usepackage{titlefoot}
\usepackage{soul}
\usepackage{caption}
\usepackage{afterpage}
\usepackage{authblk}
\usepackage{makecell}
\usepackage{listings}
\usepackage{amsthm}
\usepackage{bbm}

\usepackage{cleveref}
\crefname{figure}{Fig.}{Figs.}
\crefname{table}{Table}{Tables}
\crefname{appendix}{App.}{Apps.}
\crefname{section}{\S}{\S\S}
\crefformat{section}{\S#2#1#3} 
\crefname{equation}{Eq.}{Eqs.}
\crefname{algorithm}{Alg.}{Algs.}
\crefname{algocf}{Alg.}{Algs.}
\crefname{theorem}{Thm.}{}
\crefname{lemma}{Lemma}{}
\crefname{prop}{Prop.}{}
\crefname{footnote}{Fn.}{}

\newtheorem{prop}{Proposition}

\usepackage{xcolor}
\newcommand{\mvar}{}

\newcommand{\word}{{\mvar\ensuremath{u}}\xspace}

\newcommand{\words}{{\mvar\ensuremath{\boldsymbol{u}}}\xspace}

\newcommand{\wordt}{{\mvar\ensuremath{\word_t}}\xspace}

\newcommand{\vords}{{\mvar\ensuremath{\boldsymbol{v}}}\xspace}
\newcommand{\phuman}{{\mvar {p_{\mathrm{H}}}}}
\newcommand{\punigram}{{\mvar {q_{\mathrm{H}}}}}
\newcommand{\punigramopt}{{\mvar {q_{\mathrm{H}}}}}

\newcommand{\alphabet}{{\mvar\ensuremath{\Sigma}}\xspace}
\newcommand{\eos}{{\mvar\ensuremath{\textsc{eos}}}\xspace}
\newcommand{\eosalphabet}{{\mvar\ensuremath{\overline{\alphabet}}}\xspace}
\newcommand{\emptystr}{{\mvar \varepsilon}}

\newcommand{\corpus}{{\mvar\ensuremath{\mathcal C}}\xspace}

\newcommand{\rt}{{\mvar\ensuremath{r}}\xspace}
\newcommand{\rtn}{{\mvar\ensuremath{\rt_n}}\xspace}
\newcommand{\rtpred}{{\mvar\ensuremath{ \widehat{\rt} }}\xspace}

\newcommand{\params}{{\mvar\ensuremath{ \boldsymbol{\beta} }}\xspace}
\newcommand{\paramsbase}{{\mvar\ensuremath{ \boldsymbol{\beta}^{\intercal}_{\text{b}}}}\xspace}

\newcommand{\func}{{\mvar\ensuremath{ f_{\params} }}\xspace}
\newcommand{\dll}{{\mvar\ensuremath{ \Delta_{\mathrm{llh}}}\xspace}}

\newcommand{\pmi}{{\mvar\ensuremath{\mu_{\mathrm{H}}}\xspace}}
\newcommand{\unigramsurprisal}{{\mvar\ensuremath{\upsilon_{\mathrm{H}}}}}
\newcommand{\prefixhuman}{{\mvar\ensuremath{\pi_{\mathrm{H}}}\xspace}}
\newcommand{\samplemean}{{\mvar\ensuremath{\hat{\mu}}\xspace}}
\newcommand{\counts}{{\mvar \#\xspace}}
\newcommand{\KL}{{\mvar\mathrm{KL}}}

\newcommand{\ctx}{{\mvar \boldsymbol{c}}}
\newcommand{\unit}{{\mvar u}}
\newcommand{\eosunit}{{\mvar \overline{\unit}}}
\newcommand{\normconst}{{\mvar Z}}
\newcommand{\normconstprefix}{{\mvar \normconst_\prefixhuman}}
\newcommand{\normconstunigram}{{\mvar \normconst_\punigram}}
\DeclareMathOperator*{\argmin}{{\mvar \mathrm{argmin}}}

\newcommand{\bsurp}{{\mvar\ensuremath{ \beta_{\surprisal} }}\xspace}
\newcommand{\bfreq}{{\mvar\ensuremath{ \beta_{\unigramsurprisal} }}\xspace}

\newcommand{\bzero}{{\mvar\ensuremath{ \beta_{0} }}\xspace}
\newcommand{\pred}{{\mvar\ensuremath{ \mathbf{X}}}\xspace}

\newcommand{\predbase}{{\mvar\ensuremath{ \pred_{\text{b}}}}\xspace}

\newcommand{\defeq}[0]{\mathrel{\stackrel{\textnormal{\tiny def}}{=}}}

\newcommand{\prob}{\mathbb{P}}
\newcommand{\measure}{{\mvar \prefixhuman\cdot\phuman}}
\newcommand{\power}{\mathcal{P}}
\newcommand{\surprisalrv}{{\mvar \mathbf{I}_{\mathrm{H}}}}
\newcommand{\surprisal}{{\mvar \iota_{\mathrm{H}}}}
\newcommand{\frequencyrv}{{\mvar \mathbf{Y}_{\mathrm{H}}}}
\newcommand{\frequencyrvcomp}{{\mvar \mathbf{Y}^{\perp}_{\mathrm{H}}}}
\newcommand{\pmirv}{{\mvar \mathbf{M}_{\mathrm{H}}}}
\newcommand{\hilbertspace}{{\mvar\mathcal{H}}}
\newcommand{\hilbertsubset}{{\mvar\mathcal{C}}}
\newcommand{\expectation}{\mathbb{E}}
\newcommand{\cov}{{\mvar\ensuremath{\mathrm{Cov}}}}

\newcommand{\proj}{\mathrm{proj}}
\newcommand{\reals}{{\mvar\ensuremath{\mathbb{R}}}}

\newcommand{\RD}{{\mvar\ensuremath{\mathbb{R}^D}}}

\newcommand{\genericrv}{{\mvar \mathbf X}}
\newcommand{\genericrvy}{{\mvar \mathbf Y}}
\newcommand{\genericrvz}{{\mvar \mathbf Z}}
\newcommand{\norm}[1]{{\mvar ||#1||_{\hilbertspace}}}
\newcommand{\normsq}[1]{{\mvar ||#1||^2_{\hilbertspace}}}

\newcommand{\distance}{{\mvar d}}

\newcommand{\nsamples}{{\mvar N}}
\newcommand{\realsn}{{\mvar \reals^\nsamples}}
\newcommand{\realsntwo}{{\mvar \reals^{\nsamples \times 2}}}
\newcommand{\predmatrix}{{\mvar \mathbf{X}}}
\newcommand{\responsevar}{{\mvar \mathbf{y}}}
\newcommand{\predone}{{\mvar \mathbf{x}_1}}
\newcommand{\predtwo}{{\mvar \mathbf{x}_2}}
\newcommand{\predresid}{{\mvar \mathbf{x}_{1\text{-res}}}}
\newcommand{\paramestzero}{{\mvar\ensuremath{ \hat{\beta}^{\text{ols}}_0 }}\xspace}
\newcommand{\paramestone}{{\mvar\ensuremath{ \hat{\beta}^{\text{ols}}_1 }}\xspace}

\newcommand{\defn}[1]{\textbf{#1}}
 
\usepackage[textsize=tiny,textwidth=0.85in]{todonotes}

\definecolor{darkpurple}{rgb}{0.5,0.2,0.8}

\newcommand{\ethz}{1}
\newcommand{\cls}{2}
\newcommand{\uzh}{3}

\title{On the Role of Context in Reading Time Prediction}

\author{\textbf{Andreas Opedal}$^{\ethz,\cls}$~\;~\;~\textbf{Eleanor Chodroff}$\,^{\uzh}$~\;~\;~\textbf{Ryan Cotterell}$^{\ethz}$~\;~\;~\textbf{Ethan Gotlieb Wilcox}$^{\ethz}$ \\
$^{\ethz}$ETH Z{\"u}rich
    \quad $^{\cls}$Max Planck ETH Center for Learning Systems
    \quad $^{\uzh}$University of Z{\"u}rich\\
    \texttt{\{\href{andreas.opedal@inf.ethz.ch}{andreas.opedal},\href{ryan.cotterell@inf.ethz.ch}{ryan.cotterell},\href{ethan.wilcox@inf.ethz.ch}{ethan.wilcox}\}@inf.ethz.ch} \\
    \texttt{\href{eleanor.chodroff@uzh.ch}{eleanor.chodroff@uzh.ch}} 
}

\begin{document}
\maketitle
\begin{abstract}
We present a new perspective on how readers integrate context during real-time language comprehension. Our proposals build on surprisal theory, which posits that the processing effort of a linguistic unit (e.g., a word) is an affine function of its in-context information content. We first observe that surprisal is only one out of many potential ways that a contextual predictor can be derived from a language model. Another one is the pointwise mutual information (PMI) between a unit and its context, which turns out to yield the same predictive power as surprisal when controlling for unigram frequency. Moreover, both PMI and surprisal are correlated with frequency. This means that neither PMI nor surprisal contains information about context alone. In response to this, we propose a technique where we project surprisal onto the orthogonal complement of frequency, yielding a new contextual predictor that is uncorrelated with frequency. Our experiments show that the proportion of variance in reading times explained by context is a lot smaller when context is represented by the orthogonalized predictor. From an interpretability standpoint, this indicates that previous studies may have overstated the role that context has in predicting reading times.

\vspace{.2em}
\hspace{1.25em}\includegraphics[width=1.25em,height=1.25em]{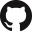}{\hspace{.75em}\parbox{\dimexpr\linewidth-2\fboxsep-2\fboxrule}{\url{https://github.com/rycolab/context-reading-time}}}

\end{abstract}

\section{Introduction}

Surprisal theory \citep{hale-2001-probabilistic, levy2008expectation} posits that the amount of effort it takes to process a linguistic unit is an affine function of its in-context information content, i.e., its surprisal. 
Numerous studies have found empirical support for surprisal theory across different reading measurement methods, languages, and language models \citep{smith2013effect, wilcox-etal:2020-on-the-predictive-power, kuribayashi2021lower, meister2021revisiting, wilcox-2023-testing, shainetal24}, particularly when controlling for additional effects such as frequency. 
In this work, we take a critical look at surprisal theory as an adequate explanation for the role of context in reading time prediction, starting from a simple observation: Surprisal is but one quantity that can be derived from a language model to represent the effect of context \citep{giulianelli2024generalized}. 
We first show that, as an alternative to surprisal, one could take an association-based view on real-time language comprehension and model it as a function of the pointwise mutual information (PMI) between a unit and its context. 
Because PMI, surprisal, and frequency are collinear, all linear models with just two of these covariates are equivalent in terms of their predictive power.
This simple identity therefore implies that all empirical validation of surprisal theory based on linear regression modeling also lends support for an association-based theory of language processing.

This raises the question of whether there is a more suitable way to estimate the effect that context has on reading time.
We argue that, given that frequency is known to play an important role in processing effort \citep{Broadbent1967WordfrequencyEA, Inhoff1986, Rayner1986, bybee2006frequency}, a more interesting construct to analyze should be what context contributes \emph{beyond} what is already captured by frequency. 
To obtain a predictor that represents just that, we propose a technique where we project surprisal onto the orthogonal complement of frequency, ensuring that they are uncorrelated. 
This process effectively disentangles the contextual and non-contextual information into different covariates in our regressions and closely resembles residualization.\footnote{See \cref{sec:residualization} for more discussion on residualization.\looseness=-1}

To test whether the choice of contextual predictor matters empirically, we measure how much the variance in reading times explained by the contextual predictor changes when substituting surprisal for the orthogonalized context predictor. 
We find that our proposed predictor results in much smaller explained variance.
Our results suggest that empirical work on surprisal theory has \emph{over}estimated the effect that context has on reading times.

\section{Predictive Models of Reading Behavior}

We seek to model the cognitive processing difficulty of a \defn{unit} $\unit$, e.g., a word, drawn from an \defn{alphabet} $\alphabet$. 
Additionally, we augment $\alphabet$ to include a unique $\eos \not\in \alphabet$ symbol which indicates the end of an utterance; we further define $\eosalphabet \defeq \alphabet \cup \{\eos\}$. 
Let $\alphabet^*$ be the set of all strings over the alphabet $\alphabet$; we write $\words \in \alphabet^*$ for a string, $\wordt$ for the $t^{\text{th}}$ unit in $\words$, $|\words|$ for the number of units in $\words$, and $\words \vords$ for the concatenation of $\words$ with another string $\vords$. 
Given a string $\ctx \unit$, we are interested in how $\unit$'s processing effort is shaped by its \defn{context} of preceding units $\ctx$.

A common psychometric proxy for the cognitive processing difficulty of $\unit$ is the time it takes a human to read $\unit$, typically, as measured in an eye-tracking study \cite{Rayner1998EyeMI}.
In general terms, we are interested in empirically assessing some theory of cognitive processing difficulty, which can be thought of as a collection of unit-level properties that are implicated in determining processing effort as measured by reading times.
The most common type of evidence adduced to support such theories comes from (generalized) linear modeling.
We define a \defn{predictor function} as a function of type $\pred \colon \alphabet^* \times \eosalphabet \rightarrow  \RD$, i.e., a function that maps a context--unit pair to a $D$-dimensional real vector.
We model the reading time measurements as a linear model $\func$ conditioned on $\pred(\ctx, \eosunit)$, i.e., $\rt(\ctx, \eosunit) \sim \func\left(\;\cdot \mid \pred(\ctx, \eosunit)\right)$ where $\params \in \RD$ is a real-valued parameter vector. 
A model whose expected value, $\rtpred(\ctx, \eosunit) = \params^{\intercal}\pred(\ctx, \eosunit)$, achieves high likelihood on held-out data lends empirical support to the theory that the factors measured by the predictors in $\pred$ underlie the process of reading. \looseness=-1

\subsection{Language Modeling Background}

We are particularly interested in predictors that are derived from language models (LMs) $\phuman$, which are distributions over $\alphabet^*$.\footnote{The subscript $\cdot_{\mathrm{H}}$ suggests that $\phuman$ is the human LM.
}
A relevant construct is the probability distribution over prefixes $\ctx \in \alphabet^*$, 
called \defn{normalized prefix probability}:
\begin{equation}
 \prefixhuman(\ctx) = \frac{1}{\normconstprefix} \sum_{\words \in \alphabet^*} \phuman(\ctx \words),
\end{equation}
where the normalizing constant $\normconstprefix$ is
\begin{align}\label{eq:norm-prefix}
\normconstprefix &= 
1 + \sum_{\words \in \alphabet^*} \phuman(\words) |\words|.  
\end{align}
See \cref{prop:prefix-normalizing} in \cref{sec:prefix-normalizing} for a proof of \cref{eq:norm-prefix}.
In words, this identity says that the normalized prefix probability exists when the expected string length is finite, which is the case for transformer-based LMs \citep{du-etal-2023-measure}. 
For simplicity, we further assume that $\phuman(\words) > 0$ for all $\words \in \alphabet^*$.
This assumption holds true in practice due to the softmax function \citep{boltzmann1868studien, gibbs1902}, which enforces the probability estimates to be strictly positive.
Then, for all $\unit \in \alphabet$,\looseness=-1
\begin{equation}\label{eq:autoregressive}
    \phuman(\unit \mid \ctx) \defeq \frac{\prefixhuman(\ctx \unit)}{\prefixhuman(\ctx)}.
\end{equation}
The $\eos$ symbol is special in the sense that
\begin{equation}\label{eq:autoregressive-eos}
    \phuman(\eos \mid \ctx) \defeq \frac{\phuman(\ctx)}{\prefixhuman(\ctx)}.
\end{equation}
Thus, $\phuman(\eosunit \mid \ctx)$ is a probability distribution over $\eosalphabet$. Importantly, note that $\phuman(\eosunit \mid \ctx)$ is \emph{not} the probability of $\ctx\eosunit$ as an entire string given that we know $\ctx$, only that $\eosunit$ follows $\ctx$. 

\subsection{Frequency as a Predictor of Reading Time}\label{sec:unigram-surprisal}

Previous studies (e.g., \citealp{shain-2019-large, shain2024dissociate}) have investigated the effect of frequency, operationalized as \defn{unigram surprisal}, on reading time. 
A unigram LM $\punigram$ is a distribution over $\alphabet^*$ where, when a string is sampled autoregressively, each unit is conditionally independent of the context. 
In notation, we write $\punigram(\eosunit)$ for the probability of $\eosunit$ independent of context.\looseness=-1

We now consider the unigram model that \emph{best approximates} the human LM $\phuman$ in the sense of the forward Kullback--Leibler divergence $\KL(\phuman \mid\mid q)$.
We can compute the minimizer $\punigram$ in closed form.
We define the following function that counts the number of occurrences of a unit $\eosunit\in\eosalphabet$ in $\words$:\looseness=-1
\begin{equation}
    \counts(\words, \eosunit) \defeq \sum\limits_{t=1}^{|\words|}  \mathbbm 1{\{ \eosunit = \wordt\}} + \mathbbm 1{\{ \eosunit = \eos\}}.
\end{equation}
Then, the minimizing unigram LM, factored autoregressively, is given by
\begin{align}\label{eq:unigram-surprisal}
\punigramopt(\eosunit) = \frac{1}{\normconstunigram} \sum\limits_{\words \in \alphabet^*} \phuman(\words) \counts(\words, \eosunit),
\end{align}
where the normalizing constant $\normconstunigram$ is necessarily finite for language models of finite expected length.\footnote{As a simple counterexample, consider an LM $\phuman$ with $\alphabet=\{a\}$ and $\phuman(a^n)=\frac{1}{\pi^2/6} \cdot \frac{1}{n^2}$ for $n\in\mathbb Z_{>0}$, i.e., where the probabilities are globally normalized by $\frac{\pi^2}{6}$, the solution to the Basel problem. The expected count of $a$ would depend on $\sum_{n=1}^{\infty} \frac{1}{n^2}\cdot n$, which is divergent. Thus, $\normconstunigram = \infty$.
} Then, given an LM $\phuman$ with a unigram LM $\punigramopt$, the unigram surprisal is given by
\begin{align}
     \unigramsurprisal(\eosunit) \defeq -\log \punigramopt(\eosunit).
\end{align}
We will refer to unigram surprisal as \defn{frequency} for the remainder of this paper. Importantly, frequency is often considered as a control variable, rather than the factor being investigated in support of a particular cognitive theory of language processing.\footnote{Additional common controls include unit length (as measured, e.g., by its orthographic representation), as well as length and frequency of previous units. The latter are included to account for spillover effects, where reading-time slowdowns triggered by a particular unit appear after a time delay.} 

\subsection{Surprisal as a Predictor of Reading Time
}

A common claim is that reading is mediated by \defn{contextual surprisal} \citep{shannon1948}, defined as
\begin{equation} \label{eq:surprisal}
    \surprisal(\ctx, \eosunit) \defeq - \log \phuman(\eosunit \mid \ctx).
\end{equation}
Indeed, this claim has received much empirical support \citep[\emph{inter alia}]{hale-2001-probabilistic, demberg-keller-2008, Smith2008-SMIOPT-2}. 
Importantly, there is evidence that the particular functional relationship, called the \defn{linking function}, between contextual surprisal and reading time is affine\footnote{Previous work often refer to this affine function as \emph{linear}.} \citep{smith2013effect, wilcox-2023-testing, shainetal24}, justifying the use of linear regression modeling. \looseness=-1

\subsection{PMI as a Predictor of Reading Time}\label{sec:pmi}

Next, we point out an alternative way of deriving a contextual predictor from an LM, namely, as the \defn{pointwise mutual information} (\defn{PMI}; \citealp{fano1961transmission}) between a unit and its context. PMI measures association, 
and has been an important notion in NLP \citep{church-hanks-1990-word, NIPS2014_feab05aa} and, more recently, psycholinguistics \citep{tsipidi2024curves,wilcox2024regressions}. The PMI between a unit $\eosunit \in \eosalphabet$ and its context $\ctx \in \alphabet^*$ is\looseness=-1
\begin{align}
    \pmi(\ctx, \eosunit) \defeq \log \frac{\phuman(\eosunit \mid \ctx) \prefixhuman(\ctx)}{\prefixhuman(\ctx) \punigram(\eosunit)}.
\end{align}
The probability that $\ctx$ and $\eosunit$ occur together is expressed in the numerator (rewritten using \cref{eq:autoregressive,eq:autoregressive-eos}). The denominator expresses what this probability would be if $\ctx$ and $\eosunit$ were independent. 

If PMI is predictive of reading times, then that would suggest a theory positing that the \emph{strength of association} that the observed unit has with its context is part of what determines the effort it takes to process it.
It turns out that many of the empirical results that have been published in support of surprisal theory, actually, by courtesy of the assumed affine linking function, provide an equal amount of evidence for a PMI-based theory.
To see this, first note that we can rewrite PMI as the difference between frequency and surprisal:
\begin{subequations}
\begin{align}
     \pmi(&\ctx, \eosunit) = \log \frac{\phuman(\eosunit \mid \ctx)}{\punigram(\eosunit) } \\
    & \quad = \unigramsurprisal(\eosunit) - \surprisal(\ctx, \eosunit). \label{eq:pmi-decomposition}
\end{align}
\end{subequations}
This equation shows that $\unigramsurprisal$, $\surprisal$ and $\pmi$ are linearly dependent in a certain Hilbert space, which we will introduce in \cref{sec:disentangling}.
Now, under a linear model $\func$ with \emph{only} surprisal and frequency as predictors, the expected value of $\func$, denoted by $\rtpred(\ctx, \eosunit)$, is given by\looseness=-1
\begin{align}
    \rtpred(&\ctx, \eosunit) =  \bzero + \bfreq \unigramsurprisal(\eosunit) + \bsurp \surprisal(\ctx, \eosunit).
\end{align}
By adding and subtracting an additional $\bsurp\unigramsurprisal(\eosunit)$ term, this can be rewritten as
\begin{align} \label{eq:surp-pmi-equiv}
    \rtpred(&\ctx, \eosunit) =  \bzero + (\bfreq+\bsurp ) \unigramsurprisal( \eosunit) - \bsurp \pmi(\ctx, \eosunit).
\end{align}
(We suppressed an intermediate step, given in \Cref{app:pmi-rewrite}.)
Thus, it turns out that the very same coefficient that is typically taken to indicate the effect of surprisal also has an alternative interpretation as the negative effect of PMI. Furthermore, the predictive power of a linear model with surprisal and frequency is \emph{the same} as that of a linear model with PMI and frequency. In other words, if frequency is provided as a predictor, additionally adding surprisal as a predictor is no more predictive than adding PMI \textit{ceteris paribus}. 
However, two such models will differ in the coefficient assigned to frequency: $\bfreq$ in the surprisal and frequency model, versus $\bfreq + \bsurp$ in the PMI and frequency model. As a consequence, they will also differ in terms of the strength of the effect attributed to the predictor that stands in for context, i.e., surprisal or PMI.\looseness=-1

\section{Disentangling the Effect of Context}\label{sec:disentangling}

As there is a large and established body of work showing that frequency plays a major role in explaining the effort it takes to process words (see, e.g., \citealp{bybee2006frequency}), we argue that the interest of surprisal theory lies in understanding what \emph{additional} effect there is of contextual information beyond frequency.
The exposition above implies that neither surprisal nor PMI should receive special status as a measure of the effect of context.  
Moreover, both surprisal and PMI are correlated with frequency\footnote{For example, in the dataset we use in our studies in \Cref{sec:experiments} we observe a Pearson correlation coefficient of $r=0.57; p<0.001$ for surprisal and $r=0.52; p<0.001$ for PMI.} and all three are collinear.

We now present a simple technique to decorrelate frequency from surprisal, resulting in a new predictor that is engineered to be disentangled from frequency.
Importantly, our technique attributes the shared effect of frequency and surprisal on reading time to frequency, and then creates a new predictor which represents the effect of surprisal that is not shared with frequency. We argue that this new predictor is more relevant to study than either surprisal or PMI since it represents only the effect of context.\looseness=-1

Our technical exposition starts with an underlying probability space $( \alphabet^* \times \eosalphabet, \power( \alphabet^* \times \eosalphabet), \measure)$.
Next, consider the following random variables under this probability space: $\surprisalrv$ encoding the distribution over surprisals $\surprisal(\ctx, \eosunit)$ of the next unit given a context, $\pmirv$ encoding the distribution over PMIs $\pmi(\ctx, \eosunit)$ between the next unit and a context, and $\frequencyrv$ encoding the distribution over frequencies $\unigramsurprisal(\eosunit)$ of a unit.
Note that $\surprisalrv$, $\pmirv$ and $\frequencyrv$ are real-valued random variables
and that $\frequencyrv$ is \emph{constant} in $\ctx$.
They are elements of a Hilbert space $\hilbertspace$ over $\reals$ containing all random variables under the above probability space that have finite second moment \citep{rudin1987realcomplex}.
The inner product on $\hilbertspace$ is given by\looseness=-1
\begin{align}
 \label{eq:inner-product}
    \langle \genericrv, \genericrvy \rangle  & \defeq \expectation[\genericrv\genericrvy] \\ & = \sum_{\substack{\ctx \in \alphabet^*, \\ \eosunit \in \eosalphabet}} \prefixhuman(\ctx) \phuman(\eosunit \mid \ctx) \genericrv(\ctx, \eosunit) \genericrvy(\ctx, \eosunit). \nonumber
\end{align}
\cref{sec:hilbert-space-proof} provides further details on why $\hilbertspace$ is indeed a Hilbert space over $\reals$ with the above inner product.
With $\hilbertspace$ being a Hilbert space, we can take projections on $\hilbertspace$. Taking the projection of $\surprisalrv$ onto the orthogonal complement of $\frequencyrv$ we get
\begin{equation}\label{eq:orthogonalized-predictor}
\proj_{\frequencyrvcomp}(\surprisalrv)  = \surprisalrv -  \frac{\langle \surprisalrv,  \frequencyrv \rangle}{\langle \frequencyrv, \frequencyrv \rangle } \frequencyrv.
\end{equation}
Projecting in this manner results in an orthogonalization in the sense that $\langle \frequencyrv,  \proj_{\frequencyrvcomp}(\surprisalrv) \rangle = 0$, as a consequence of the Hilbert projection theorem \citep[pp. 306-9]{rudin1991functional}. See \cref{sec:projection} for a proof.
If the expected values of at least one of the random variables $\genericrv$ and $\genericrvy$ is $0$, which can be achieved by a simple mean-centering transformation, then 
\begin{equation}
\begin{aligned}
\langle \genericrv, \genericrvy \rangle &= \expectation[\genericrv \genericrvy] \\
&= \cov(\genericrv, \genericrvy) + \underbrace{\expectation[\genericrv]\expectation[\genericrvy]}_{=0}.
\end{aligned}
\end{equation}
Thus, if $\frequencyrv$ and $\surprisalrv$ are mean-centered the covariance between $\frequencyrv$ and $\proj_{\frequencyrvcomp}(\surprisalrv)$ will be $0$, i.e., they will be decorrelated.
The random variable $\proj_{\frequencyrvcomp}(\surprisalrv)$ constitutes a new predictor variable, which we term \defn{orthogonalized surprisal}.
In words, orthogonalized surprisal represents the effect of context and is disentangled from frequency.\footnote{The method presented can in principle be applied to \emph{any} pair of predictor variables that live in $\hilbertspace$, e.g., taking $\pmirv$ instead of $\surprisalrv$ yields an equivalent context predictor. 
We also consider orthogonalized unit length in our experiments in \cref{sec:experiments}.
\label{fn:length}}

\section{Variance Explained by Context}\label{sec:experiments}

\begin{figure*}[t]
    \centering
    \vspace{-2pt}
    \includegraphics[width=0.95\textwidth]{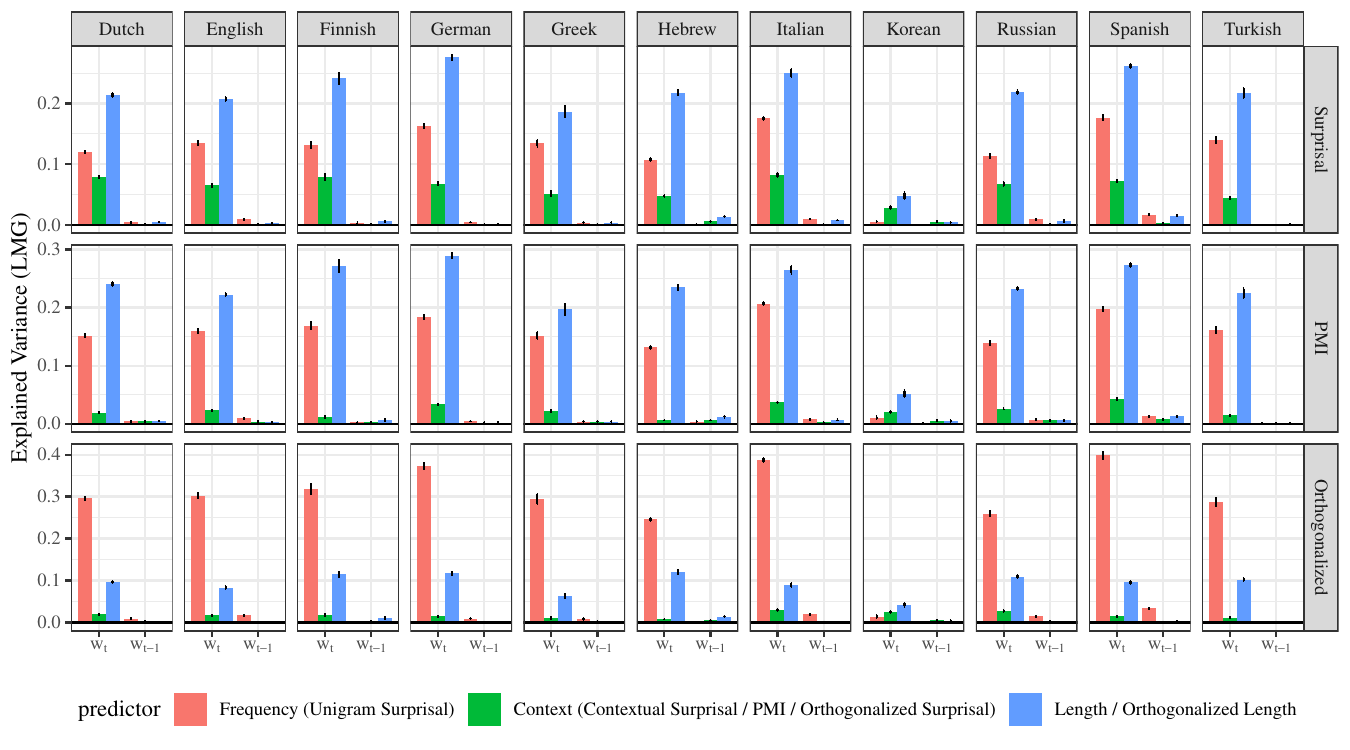}
    \vspace{-10pt}
    \caption{Proportion of total variance explained by the predictors \citep{kruskal1987relative}, across languages and linear regression models. 
    Summing the values represented by the four bars yields the coefficient of determination $R^2$. 
    Note that the $R^2$ values are the same across the three models, as a consequence of the collinearity discussed in \cref{sec:pmi}. Error bars are 95\% confidence intervals across folds of data. We observe only a small proportion of explained variance by context when excluding the variance it shares with frequency, as in the orthogonalized surprisal predictor.
    }
    \label{fig:results-exp2}
  \vspace{-6.0pt}
\end{figure*}

We now seek an empirical understanding of how the new orthogonalized surprisal predictor influences the importance attributed to context in reading time prediction. In addition to the experiment presented here, we also compared the predictive power for \emph{nonlinear} models across different predictors. Those experiments are discussed in \cref{sec:additional-experiments}.

\paragraph{Dataset and Predictors.} We use gaze duration times from the Multilingual Eye-movement Corpus (MECO; \citealp{siegelman2022expanding}), which consists of word-level eye-tracking measurements from several languages.
To obtain surprisal estimates, we approximate $\phuman$ with mGPT \citep{mgpt2024}, which is a multilingual LM based on GPT-2. 
The frequency estimates are from \citet{robyn_speer_2022_7199437}\footnote{See \citet{nikkarinen-etal-2021-modeling} for a more nuanced estimation technique.} and PMI is computed through the decomposition given in \cref{eq:pmi-decomposition}.
Orthogonalized surprisal is obtained by approximating $\langle \surprisalrv,  \frequencyrv \rangle$ and $\langle \frequencyrv, \frequencyrv \rangle$ using only the words that occur in a training corpus. We transform surprisal and frequency to be mean-centered, and use the transformed data to compute unbiased sample variances and covariances. 
Those sample estimates are then plugged into \cref{eq:orthogonalized-predictor}. \cref{sec:orthogonalized-linear} gives the formulaic details of this approximation. We also include a word length predictor. 
Word length is correlated with frequency \citep{Zipf49}, so we orthogonalize it in the same manner as surprisal. \Cref{app:methods} gives more details on the dataset and predictor variables.

\paragraph{Experimental Setup.} We fit three linear models using ordinary least squares, with predictors: (i) surprisal, frequency, and length, (ii) PMI, frequency, and length, and (iii) orthogonalized surprisal, frequency, and orthogonalized length.
(An alternative perspective would be to use orthogonalized \emph{frequency}, which assumes that \textit{surprisal} is the more fundamental predictor. We provide such results, which are are consistent with the current conclusions, in \cref{sec:orthogonal-frequency}.)
We include spillover effects from the previous word and fit models over ten folds of cross-validation. We then measure the relative importance of predictors by averaging over the proportion of variance explained by the predictors across orderings in which they are added to the model \citep{lmg1980, kruskal1987relative}, a technique known as LMG \citep{gromping2007relative}.\footnote{The intuition behind LMG is simple: Given a response variable $\responsevar$ and two predictors $\predone$ and $\predtwo$, the proportion of the variance of $\responsevar$ explained by $\predone$ can be taken to either depend on the correlation between $\predone$ and $\responsevar$ (if $\predone$ is added to the model \emph{before} $\predtwo$), or to depend on the partial correlation between $\predone$ and $\responsevar$ when controlling for $\predtwo$ (if $\predone$ is added to the model \emph{after} $\predtwo$). For a model with $p$ predictors, the LMG averages the explained variance over all $p!$ such orderings.} One advantage of this technique, compared to previous methods (e.g., delta log-likelihood; \citealp{goodkind-bicknell-2018-predictive}) is that LMG gives a more interpretable absolute score for predictive power. 

\paragraph{Results.} The LMG values for the different predictors across these models are visualized in \cref{fig:results-exp2}. Comparing the plots on the bottom row to the plots on the top row, we observe that the explained variance for orthogonalized surprisal is much lower than for standard contextual surprisal. These results are consistent across languages and suggest that using surprisal overestimates the importance that context has on reading time. 
We observe values for PMI that for most languages lie between those for surprisal and orthogonalized surprisal, indicating that the extent to which that PMI overestimates the context effect is smaller compared to surprisal. 
Moreover, we note that the context-based predictors are considerably less predictive than frequency and (orthogonalized) length.\footnote{Our results may help explain why surprisal estimates from larger LMs provide poorer fits to reading than those from medium-sized LMs \citep{oh2022does}; see \cref{sec:discussion}.} This holds for all languages except Korean.
Excluding Korean, we observe that the average $R^2$ values of the models across LMG orderings, i.e., the sums of the bars within each facet, range between $0.25$--$0.45$, indicating that the linear models capture a moderate proportion of the variance in gaze duration.

\section{Conclusion}
This article discusses predictors that capture how the processing effort of a unit is shaped by its context. We made the observation that there exist alternatives to the widely used surprisal predictor. Surprisal is correlated with non-contextual frequency,
so we provided a technique to disentangle contextual and non-contextual information in language models. In so doing, we found that the effect that context has on reading times appears to be small in comparison to non-contextual frequency. \looseness=-1

\section*{Limitations}

Our approach takes one predictor to remain untouched (i.e., frequency), and modifies others to reflect effects that are disassociated from the first.
As suggested above, it would thus be natural to ask what happens in the alternative setting, where surprisal remains untouched and frequency is projected onto the orthogonal complement of surprisal.  
We provide such an analysis in \cref{sec:orthogonal-frequency}. It turns out that even when attributing the shared effect of frequency and surprisal on reading times to surprisal---which is the case when replacing frequency by an orthogonalized frequency predictor---the variance explained by the frequency predictor is still higher for most languages in comparison to the surprisal predictor. This gives further support to our conclusion that context appears to play a smaller role in reading time prediction.

Furthermore, our presentation of ideas and discussion largely ignores effects of word lengths. We find word lengths to explain the most variance in reading times in our surprisal and PMI models. In addition, after residualizing word length against frequency, we find length to be the second strongest predictor, with an explained variance ranging from around $7
\%$--$12\%$.
One hypothesis is that readers may make multiple saccades within first passes of longer words, and the time it takes to plan and execute these saccades could be the underlying reason why orthogonalized word length remains explanatory even after residualization. Future work could control for this by adding in the number of saccades within a word as an additional predictor into models.\looseness=-1

We are unaware of any efficient algorithm to compute $\langle \surprisalrv,  \frequencyrv \rangle$ and $\langle \frequencyrv, \frequencyrv \rangle$ exactly, so in practical settings we must rely on estimation. Thus, it may be that the orthogonalized surprisal predictor is only ``close to'' being orthogonal to frequency in practice. The manner in which estimation is performed makes our technique similar to residualization---see \cref{sec:residualization} for a discussion. Moreover, our method only provides guarantees for predictor variables that live in $\hilbertspace$. 

Importantly, an off-by-one issue was detected in the trial, sentence and word (interest area) ID scheme for a handful of tokens in MECO (Versions 1.1 and 1.2). This was corrected for our analysis, but should be taken into consideration by other researchers using this dataset. The code to correct the data is included in our repository.
In addition, we also identified a few instances of repeated words within a sentence (e.g., ``als als''), but given their consistency across participants, we assumed these were typos. These were retained in the analysis, but could have resulted in atypical responses.

In addition, the current paper makes use of a fixed-effects linear regression model with averaged data, as opposed to the more standard mixed-effects regression. Estimation of $R^2$ values from mixed-effects models can differ depending on the researcher assumptions and has historically been under-reported due to this limitation \cite{nakagawa2013general}. Nevertheless, some proposals have been made regarding best practices \citep{nakagawa2013general, rights2018framework}. Future research should investigate the feasibility of our approach, particularly with the use of partial effect sizes (i.e., the LMG approach), but using mixed-effects models. 

Another limitation of this work is that, while we investigate several languages, these are still biased towards Indo-European languages. For example, we present results from one language only for Fino-Uralic, Semitic, Turkic, and Koreanic language families, but seven Indo-European languages. Expanding these results to even more languages would further broaden the impact of this work. In addition, we observe somewhat unique effects for Korean, where frequency accounts for a lower proportion of the variance, and length and context account for higher proportions. One possible reason for this is the Korean script (Hangul), which combines features of both alphabetic and syllabic writing systems. Future work should conduct similar analyses on different Korean datasets to determine whether this trend is a property of Korean, or just our particular Korean language dataset.

\section*{Ethical Considerations}

This work uses previously collected human data from the MECO dataset. Please see the paper that introduces this dataset \citep{siegelman2022expanding} for information about the data collection procedure. The authors foresee no ethical problems arising from the work presented here.

\section*{Acknowledgments}
We thank Cory Shain for useful discussion. Andreas Opedal acknowledges funding from the Max Planck ETH Center for Learning Systems.

\bibliography{custom}

\begin{thebibliography}{53}
\providecommand{\natexlab}[1]{#1}

\bibitem[{Boltzmann(1868)}]{boltzmann1868studien}
Ludwig Boltzmann. 1868.
\newblock \href {https://books.google.ch/books?id=vxWqmgEACAAJ} {\emph{Studien {\"u}ber das Gleichgewicht der lebendigen Kraft zwischen bewegten materiellen Punkten}}.
\newblock K.K. Hof und Staatsdruckerei.

\bibitem[{Breaugh(2006)}]{Breaugh2006}
James~A. Breaugh. 2006.
\newblock \href {https://doi.org/10.1007/s10869-005-9009-y} {Rethinking the control of nuisance variables in theory testing}.
\newblock \emph{Journal of Business and Psychology}, 20(3):429--443.

\bibitem[{Broadbent(1967)}]{Broadbent1967WordfrequencyEA}
Donald~E. Broadbent. 1967.
\newblock \href {https://api.semanticscholar.org/CorpusID:34321401} {Word-frequency effect and response bias.}
\newblock \emph{Psychological Review}, 74 1:1--15.

\bibitem[{Bybee(2006)}]{bybee2006frequency}
Joan Bybee. 2006.
\newblock \href {https://books.google.ch/books?hl=en&lr=&id=kBtREAAAQBAJ&oi=fnd&pg=PR7&ots=ruRcHedMeH&sig=nt1WpDRu7ZdARnle_EfkBsv8PcY&redir_esc=y#v=onepage&q&f=false} {\emph{Frequency of {U}se and the {O}rganization of {L}anguage}}.
\newblock Oxford University Press.

\bibitem[{Church and Hanks(1990)}]{church-hanks-1990-word}
Kenneth~Ward Church and Patrick Hanks. 1990.
\newblock \href {https://aclanthology.org/J90-1003} {Word association norms, mutual information, and lexicography}.
\newblock \emph{Computational Linguistics}, 16(1):22--29.

\bibitem[{Demberg and Keller(2008)}]{demberg-keller-2008}
Vera Demberg and Frank Keller. 2008.
\newblock \href {https://doi.org/10.1016/j.cognition.2008.07.008} {Data from eye-tracking corpora as evidence for theories of syntactic processing complexity}.
\newblock \emph{Cognition}, 109(2):193--210.

\bibitem[{Du et~al.(2023)Du, Torroba~Hennigen, Pimentel, Meister, Eisner, and Cotterell}]{du-etal-2023-measure}
Li~Du, Lucas Torroba~Hennigen, Tiago Pimentel, Clara Meister, Jason Eisner, and Ryan Cotterell. 2023.
\newblock \href {https://doi.org/10.18653/v1/2023.acl-long.543} {A measure-theoretic characterization of tight language models}.
\newblock In \emph{Proceedings of the 61st Annual Meeting of the Association for Computational Linguistics (Volume 1: Long Papers)}, pages 9744--9770, Toronto, Canada. Association for Computational Linguistics.

\bibitem[{Fano(1961)}]{fano1961transmission}
Robert~M. Fano. 1961.
\newblock \href {https://books.google.ch/books?id=2PMbtQEACAAJ} {\emph{Transmission of Information: A Statistical Theory of Communication}}.
\newblock MIT Press Classics. MIT Press.

\bibitem[{Futrell et~al.(2020)Futrell, Gibson, and Levy}]{futrell2020lossy}
Richard Futrell, Edward Gibson, and Roger~P. Levy. 2020.
\newblock \href {https://doi.org/10.1111/cogs.12814} {Lossy-context surprisal: An information-theoretic model of memory effects in sentence processing}.
\newblock \emph{Cognitive Science}, 44(3):e12814.

\bibitem[{Garc{\'\i}a et~al.(2019)Garc{\'\i}a, Salmer{\'o}n, Garc{\'\i}a, and Garc{\'\i}a}]{Garcia2019-om}
Catalina~B. Garc{\'\i}a, Rom{\'a}n Salmer{\'o}n, Claudia Garc{\'\i}a, and Jos{\'e} Garc{\'\i}a. 2019.
\newblock \href {https://www.ncbi.nlm.nih.gov/pmc/articles/PMC9041907/} {Residualization: justification, properties and application}.
\newblock \emph{Journal of Applied Statistics}, 47(11):1990--2010.

\bibitem[{Gibbs(1902)}]{gibbs1902}
Josiah~Willard Gibbs. 1902.
\newblock \href {https://www.scirp.org/reference/referencespapers?referenceid=1519525} {\emph{Elementary Principles in Statistical Mechanics}}.
\newblock Charles Scribner's Sons.

\bibitem[{Giulianelli et~al.(2024)Giulianelli, Opedal, and Cotterell}]{giulianelli2024generalized}
Mario Giulianelli, Andreas Opedal, and Ryan Cotterell. 2024.
\newblock \href {https://arxiv.org/abs/2409.10728} {Generalized measures of anticipation and responsivity in online language processing}.
\newblock In \emph{Findings of the Association for Computational Linguistics: EMNLP 2024}, Miami, Florida, USA. Association for Computational Linguistics.

\bibitem[{Goodkind and Bicknell(2018)}]{goodkind-bicknell-2018-predictive}
Adam Goodkind and Klinton Bicknell. 2018.
\newblock \href {https://doi.org/10.18653/v1/W18-0102} {Predictive power of word surprisal for reading times is a linear function of language model quality}.
\newblock In \emph{Proceedings of the 8th Workshop on Cognitive Modeling and Computational Linguistics ({CMCL} 2018)}, pages 10--18, Salt Lake City, Utah. Association for Computational Linguistics.

\bibitem[{Grömping(2007)}]{gromping2007relative}
Ulrike Grömping. 2007.
\newblock \href {https://doi.org/10.1198/000313007X188252} {Estimators of relative importance in linear regression based on variance decomposition}.
\newblock \emph{The American Statistician}, 61(2):139--147.

\bibitem[{Hale(2001)}]{hale-2001-probabilistic}
John Hale. 2001.
\newblock \href {https://aclanthology.org/N01-1021} {A probabilistic {E}arley parser as a psycholinguistic model}.
\newblock In \emph{Second Meeting of the North {A}merican Chapter of the Association for Computational Linguistics}.

\bibitem[{Hoover et~al.(2023)Hoover, Sonderegger, Piantadosi, and O’Donnell}]{hoover-etal-2023-plausibility}
Jacob~Louis Hoover, Morgan Sonderegger, Steven~T. Piantadosi, and Timothy~J. O’Donnell. 2023.
\newblock \href {https://doi.org/10.1162/opmi_a_00086} {{The Plausibility of Sampling as an Algorithmic Theory of Sentence Processing}}.
\newblock \emph{Open Mind}, 7:350--391.

\bibitem[{Inhoff and Rayner(1986)}]{Inhoff1986}
Albrecht~Werner Inhoff and Keith Rayner. 1986.
\newblock \href {https://doi.org/10.3758/BF03208203} {Parafoveal word processing during eye fixations in reading: Effects of word frequency}.
\newblock \emph{Perception {\&} Psychophysics}, 40(6):431--439.

\bibitem[{Jaeger(2010)}]{FLORIANJAEGER201023}
Florian Jaeger. 2010.
\newblock \href {https://doi.org/10.1016/j.cogpsych.2010.02.002} {Redundancy and reduction: Speakers manage syntactic information density}.
\newblock \emph{Cognitive Psychology}, 61(1):23--62.

\bibitem[{Kruskal(1987)}]{kruskal1987relative}
William Kruskal. 1987.
\newblock \href {https://doi.org/10.1080/00031305.1987.10475432} {Relative importance by averaging over orderings}.
\newblock \emph{The American Statistician}, 41(1):6--10.

\bibitem[{Kuperman et~al.(2008)Kuperman, Bertram, and Baayen}]{kuperman2008}
Victor Kuperman, Raymond Bertram, and R.~Harald Baayen. 2008.
\newblock \href {https://doi.org/10.1080/01690960802193688} {Morphological dynamics in compound processing}.
\newblock \emph{Language and Cognitive Processes}, 23(7-8):1089--1132.

\bibitem[{Kuribayashi et~al.(2022)Kuribayashi, Oseki, Brassard, and Inui}]{kuribayashi-etal-2022-context}
Tatsuki Kuribayashi, Yohei Oseki, Ana Brassard, and Kentaro Inui. 2022.
\newblock \href {https://doi.org/10.18653/v1/2022.emnlp-main.712} {Context limitations make neural language models more human-like}.
\newblock In \emph{Proceedings of the 2022 Conference on Empirical Methods in Natural Language Processing}, pages 10421--10436, Abu Dhabi, United Arab Emirates. Association for Computational Linguistics.

\bibitem[{Kuribayashi et~al.(2021)Kuribayashi, Oseki, Ito, Yoshida, Asahara, and Inui}]{kuribayashi2021lower}
Tatsuki Kuribayashi, Yohei Oseki, Takumi Ito, Ryo Yoshida, Masayuki Asahara, and Kentaro Inui. 2021.
\newblock \href {https://doi.org/10.18653/v1/2021.acl-long.405} {Lower perplexity is not always human-like}.
\newblock In \emph{Proceedings of the 59th Annual Meeting of the Association for Computational Linguistics and the 11th International Joint Conference on Natural Language Processing (Volume 1: Long Papers)}, pages 5203--5217, Online. Association for Computational Linguistics.

\bibitem[{Levy and Goldberg(2014)}]{NIPS2014_feab05aa}
Omer Levy and Yoav Goldberg. 2014.
\newblock \href {https://proceedings.neurips.cc/paper_files/paper/2014/file/feab05aa91085b7a8012516bc3533958-Paper.pdf} {Neural word embedding as implicit matrix factorization}.
\newblock In \emph{Advances in Neural Information Processing Systems}, volume~27. Curran Associates, Inc.

\bibitem[{Levy(2008)}]{levy2008expectation}
Roger Levy. 2008.
\newblock \href {https://doi.org/10.1016/j.cognition.2007.05.006} {Expectation-based syntactic comprehension}.
\newblock \emph{Cognition}, 106(3):1126--1177.

\bibitem[{Lindeman et~al.(1980)Lindeman, Merenda, and Gold}]{lmg1980}
Richard~H. Lindeman, Peter~F. Merenda, and Ruth~Z. Gold. 1980.
\newblock \href {http://lib.ugent.be/catalog/rug01:000474859} {\emph{Introduction to bivariate and multivariate analysis}}.
\newblock Glenview (Ill.) : Scott.

\bibitem[{Meister et~al.(2021)Meister, Pimentel, Haller, J{\"a}ger, Cotterell, and Levy}]{meister2021revisiting}
Clara Meister, Tiago Pimentel, Patrick Haller, Lena J{\"a}ger, Ryan Cotterell, and Roger Levy. 2021.
\newblock \href {https://doi.org/10.18653/v1/2021.emnlp-main.74} {Revisiting the {U}niform {I}nformation {D}ensity hypothesis}.
\newblock In \emph{Proceedings of the 2021 Conference on Empirical Methods in Natural Language Processing}, pages 963--980, Online and Punta Cana, Dominican Republic. Association for Computational Linguistics.

\bibitem[{Nakagawa and Schielzeth(2013)}]{nakagawa2013general}
Shinichi Nakagawa and Holger Schielzeth. 2013.
\newblock \href {https://besjournals.onlinelibrary.wiley.com/doi/10.1111/j.2041-210x.2012.00261.x} {A general and simple method for obtaining r2 from generalized linear mixed-effects models}.
\newblock \emph{Methods in ecology and evolution}, 4(2):133--142.

\bibitem[{Nikkarinen et~al.(2021)Nikkarinen, Pimentel, Blasi, and Cotterell}]{nikkarinen-etal-2021-modeling}
Irene Nikkarinen, Tiago Pimentel, Dami{\'a}n Blasi, and Ryan Cotterell. 2021.
\newblock \href {https://doi.org/10.18653/v1/2021.findings-acl.326} {Modeling the unigram distribution}.
\newblock In \emph{Findings of the Association for Computational Linguistics: ACL-IJCNLP 2021}, pages 3721--3729, Online. Association for Computational Linguistics.

\bibitem[{Oh and Schuler(2023{\natexlab{a}})}]{oh-schuler-2023-transformer}
Byung-Doh Oh and William Schuler. 2023{\natexlab{a}}.
\newblock \href {https://doi.org/10.18653/v1/2023.findings-emnlp.128} {Transformer-based language model surprisal predicts human reading times best with about two billion training tokens}.
\newblock In \emph{Findings of the Association for Computational Linguistics: EMNLP 2023}, pages 1915--1921, Singapore. Association for Computational Linguistics.

\bibitem[{Oh and Schuler(2023{\natexlab{b}})}]{oh2022does}
Byung-Doh Oh and William Schuler. 2023{\natexlab{b}}.
\newblock \href {https://doi.org/10.1162/tacl_a_00548} {Why does surprisal from larger transformer-based language models provide a poorer fit to human reading times?}
\newblock \emph{Transactions of the Association for Computational Linguistics}, 11:336--350.

\bibitem[{Oh et~al.(2024)Oh, Yue, and Schuler}]{oh-etal-2024-frequency}
Byung-Doh Oh, Shisen Yue, and William Schuler. 2024.
\newblock \href {https://aclanthology.org/2024.eacl-long.162} {Frequency explains the inverse correlation of large language models{'} size, training data amount, and surprisal{'}s fit to reading times}.
\newblock In \emph{Proceedings of the 18th Conference of the European Chapter of the Association for Computational Linguistics (Volume 1: Long Papers)}, pages 2644--2663, St. Julian{'}s, Malta. Association for Computational Linguistics.

\bibitem[{Raffel et~al.(2020)Raffel, Shazeer, Roberts, Lee, Narang, Matena, Zhou, Li, and Liu}]{C4-corpus-2020}
Colin Raffel, Noam Shazeer, Adam Roberts, Katherine Lee, Sharan Narang, Michael Matena, Yanqi Zhou, Wei Li, and Peter~J. Liu. 2020.
\newblock \href {http://jmlr.org/papers/v21/20-074.html} {Exploring the limits of transfer learning with a unified text-to-text transformer}.
\newblock \emph{Journal of Machine Learning Research}, 21(140):1--67.

\bibitem[{Rayner(1998)}]{Rayner1998EyeMI}
Keith Rayner. 1998.
\newblock \href {https://api.semanticscholar.org/CorpusID:3015502} {Eye movements in reading and information processing: 20 years of research.}
\newblock \emph{Psychological bulletin}, 124 3:372--422.

\bibitem[{Rayner and Duffy(1986)}]{Rayner1986}
Keith Rayner and Susan~A. Duffy. 1986.
\newblock \href {https://doi.org/10.3758/BF03197692} {Lexical complexity and fixation times in reading: Effects of word frequency, verb complexity, and lexical ambiguity}.
\newblock \emph{Memory {\&} Cognition}, 14(3):191--201.

\bibitem[{Rights and Sterba(2018)}]{rights2018framework}
Jason~D. Rights and Sonya~K. Sterba. 2018.
\newblock \href {https://pubmed.ncbi.nlm.nih.gov/28301198/} {A framework of {R}-squared measures for single-level and multilevel regression mixture models.}
\newblock \emph{Psychological Methods}, 23(3):434.

\bibitem[{Rudin(1987)}]{rudin1987realcomplex}
Walter Rudin. 1987.
\newblock \href {https://books.google.com/books/about/Real_and_Complex_Analysis.html?id=NmW7QgAACAAJ&redir_esc=y} {\emph{Real and complex analysis, 3rd ed.}}
\newblock McGraw-Hill, Inc., USA.

\bibitem[{Rudin(1991)}]{rudin1991functional}
Walter Rudin. 1991.
\newblock \href {https://books.google.se/books?id=Sh_vAAAAMAAJ} {\emph{Functional Analysis}}.
\newblock International series in pure and applied mathematics. McGraw-Hill.

\bibitem[{Shain(2019)}]{shain-2019-large}
Cory Shain. 2019.
\newblock \href {https://doi.org/10.18653/v1/N19-1413} {A large-scale study of the effects of word frequency and predictability in naturalistic reading}.
\newblock In \emph{Proceedings of the 2019 Conference of the North {A}merican Chapter of the Association for Computational Linguistics: Human Language Technologies, Volume 1 (Long and Short Papers)}, pages 4086--4094, Minneapolis, Minnesota. Association for Computational Linguistics.

\bibitem[{Shain(2024)}]{shain2024dissociate}
Cory Shain. 2024.
\newblock \href {https://doi.org/10.1162/opmi_a_00119} {{Word Frequency and Predictability Dissociate in Naturalistic Reading}}.
\newblock \emph{Open Mind}, 8:177--201.

\bibitem[{Shain et~al.(2024)Shain, Meister, Pimentel, Cotterell, and Levy}]{shainetal24}
Cory Shain, Clara Meister, Tiago Pimentel, Ryan Cotterell, and Roger Levy. 2024.
\newblock \href {https://doi.org/10.1073/pnas.2307876121} {Large-scale evidence for logarithmic effects of word predictability on reading time}.
\newblock \emph{Proceedings of the National Academy of Sciences}, 121(10):e2307876121.

\bibitem[{Shannon(1948)}]{shannon1948}
Claude~E. Shannon. 1948.
\newblock \href {https://doi.org/10.1002/j.1538-7305.1948.tb01338.x} {A mathematical theory of communication}.
\newblock \emph{The Bell System Technical Journal}, 27(3):379--423.

\bibitem[{Shliazhko et~al.(2024)Shliazhko, Fenogenova, Tikhonova, Kozlova, Mikhailov, and Shavrina}]{mgpt2024}
Oleh Shliazhko, Alena Fenogenova, Maria Tikhonova, Anastasia Kozlova, Vladislav Mikhailov, and Tatiana Shavrina. 2024.
\newblock \href {https://direct.mit.edu/tacl/article/doi/10.1162/tacl_a_00633/119278/mGPT-Few-Shot-Learners-Go-Multilingual} {{mGPT: Few-Shot Learners Go Multilingual}}.
\newblock \emph{Transactions of the Association for Computational Linguistics}, 12:58--79.

\bibitem[{Siegelman et~al.(2022)Siegelman, Schroeder, Acart{\"u}rk, Ahn, Alexeeva, Amenta, Bertram, Bonandrini, Brysbaert, Chernova et~al.}]{siegelman2022expanding}
Noam Siegelman, Sascha Schroeder, Cengiz Acart{\"u}rk, Hee-Don Ahn, Svetlana Alexeeva, Simona Amenta, Raymond Bertram, Rolando Bonandrini, Marc Brysbaert, Daria Chernova, et~al. 2022.
\newblock \href {https://doi.org/https://link.springer.com/article/10.3758/s13428-021-01772-6} {Expanding horizons of cross-linguistic research on reading: The multilingual eye-movement corpus {(MECO)}}.
\newblock \emph{Behavior Research Methods}, 54(6):2843--2863.

\bibitem[{Smith and Levy(2008)}]{Smith2008-SMIOPT-2}
Nathaniel~J. Smith and Roger Levy. 2008.
\newblock \href {https://www.mit.edu/~rplevy/papers/smith-levy-2008-cogsci.pdf} {Optimal processing times in reading: {A} formal model and empirical investigation}.
\newblock In \emph{Proceedings of the 30th Annual Conference of the Cognitive Science Society}.

\bibitem[{Smith and Levy(2013)}]{smith2013effect}
Nathaniel~J. Smith and Roger Levy. 2013.
\newblock \href {https://doi.org/10.1016/j.cognition.2013.02.013} {The effect of word predictability on reading time is logarithmic}.
\newblock \emph{Cognition}, 128(3):302--319.

\bibitem[{Speer(2022)}]{robyn_speer_2022_7199437}
Robyn Speer. 2022.
\newblock \href {https://doi.org/10.5281/zenodo.7199437} {rspeer/wordfreq: v3.0}.

\bibitem[{Tsipidi et~al.(2024)Tsipidi, Nowak, Cotterell, Gotlieb~Wilcox, Giulianelli, and Warstadt}]{tsipidi2024curves}
Eleftheria Tsipidi, Franz Nowak, Ryan Cotterell, Ethan Gotlieb~Wilcox, Mario Giulianelli, and Alex Warstadt. 2024.
\newblock \href {https://aclanthology.org/2024.emnlp-main.1047/} {Surprise! {U}niform information density isn't the whole story: Predicting surprisal contours in long-form discourse}.
\newblock In \emph{Proceedings of the 2024 Conference on Empirical Methods in Natural Language Processing}, Miami, Florida, USA. Association for Computational Linguistics.

\bibitem[{Wilcox et~al.(2020)Wilcox, Gauthier, Hu, Qian, and Levy}]{wilcox-etal:2020-on-the-predictive-power}
Ethan~Gotlieb Wilcox, Jon Gauthier, Jennifer Hu, Peng Qian, and Roger~P. Levy. 2020.
\newblock \href {https://arxiv.org/abs/2006.01912} {On the predictive power of neural language models for human real-time comprehension behavior}.
\newblock In \emph{Proceedings of the 42nd Annual Meeting of the Cognitive Science Society}, page 1707–1713.

\bibitem[{Wilcox et~al.(2024)Wilcox, Pimentel, Meister, and Cotterell}]{wilcox2024regressions}
Ethan~Gotlieb Wilcox, Tiago Pimentel, Clara Meister, and Ryan Cotterell. 2024.
\newblock \href {https://doi.org/10.1016/j.cognition.2024.105765} {An information-theoretic analysis of targeted regressions during reading}.
\newblock \emph{Cognition}, 249:105765.

\bibitem[{Wilcox et~al.(2023)Wilcox, Pimentel, Meister, Cotterell, and Levy}]{wilcox-2023-testing}
Ethan~Gotlieb Wilcox, Tiago Pimentel, Clara Meister, Ryan Cotterell, and Roger~P. Levy. 2023.
\newblock \href {https://doi.org/10.1162/tacl_a_00612} {{Testing the Predictions of Surprisal Theory in 11 Languages}}.
\newblock \emph{Transactions of the Association for Computational Linguistics}, 11:1451--1470.

\bibitem[{Wurm and Fisicaro(2014)}]{wurmfisicaro2014residualizing}
Lee~H. Wurm and Sebastiano~A. Fisicaro. 2014.
\newblock \href {https://doi.org/10.1016/j.jml.2013.12.003} {What residualizing predictors in regression analyses does (and what it does not do)}.
\newblock \emph{Journal of Memory and Language}, 72:37--48.

\bibitem[{York(2012)}]{york2012residualization}
Richard York. 2012.
\newblock \href {https://doi.org/10.1016/j.ssresearch.2012.05.014} {Residualization is not the answer: Rethinking how to address multicollinearity}.
\newblock \emph{Social Science Research}, 41(6):1379--1386.

\bibitem[{Zipf(1949)}]{Zipf49}
George~K. Zipf. 1949.
\newblock \href {https://psycnet.apa.org/record/1950-00412-000} {\emph{Human Behaviour and the Principle of Least Effort}}.
\newblock Addison-Wesley.

\end{thebibliography}

\onecolumn
\appendix

\section{Normalizing the Prefix Probabilities}
\label{sec:prefix-normalizing}

Here we show that $\normconstprefix$ as defined in \cref{eq:norm-prefix} is the normalizing constant for the prefix probabilities.  

\begin{prop}\label{prop:prefix-normalizing}
    Let $\normconstprefix$ be defined as in \cref{eq:norm-prefix}, i.e., 
    \begin{equation}
        \normconstprefix = 1 + \sum_{\words \in \alphabet^*}\phuman(\words) |\words| .
    \end{equation}
    Then, $\normconstprefix$ is the normalizing constant for $\prefixhuman$, i.e., 
    \begin{equation}
        \sum_{\ctx \in \alphabet ^*} \prefixhuman(\ctx) = 1.
    \end{equation}
\end{prop}

\begin{proof}
First, note that \cref{eq:norm-prefix} can be rewritten in the following manner:
\begin{subequations}
\begin{align}
    \normconstprefix & =  1 + \sum_{\words \in \alphabet^*} \phuman(\words) |\words|  \\
    & = \sum_{\words \in \alphabet^*} \phuman(\words) + \sum_{\words \in \alphabet^*} \phuman(\words)  |\words| \\
    & = \sum_{\words \in \alphabet^*} (1 + |\words|) \phuman(\words) \\
    & = \sum_{\ctx \in \alphabet^*} \sum_{\words' \in \alphabet^*} \phuman(\ctx \words').
\end{align}
\end{subequations}
The last step follows from the fact that a string $\words$ can be segmented into two substrings $\ctx$ and $\words'$ in $1 + |\words|$ ways. 
Note the two cases where either $\ctx=\emptystr$ and $\words'=\words$, or $\ctx=\words$ and $\words'=\emptystr$ (with $\emptystr$ denoting the empty string). 
It is then easy to see that $\normconstprefix$ is a valid normalizing constant:
\begin{subequations}
\begin{align}
    \sum_{\ctx' \in \alphabet ^*} \prefixhuman(\ctx') & = \sum_{\ctx' \in \alphabet ^*} \frac{\sum_{\words \in \alphabet^*} \phuman(\ctx' \words)}{\sum_{\ctx \in \alphabet^*} \sum_{\words' \in \alphabet^*} \phuman(\ctx \words')} \\
    & = \frac{\sum_{\ctx' \in \alphabet ^*}\sum_{\words \in \alphabet^*} \phuman(\ctx' \words)}{\sum_{\ctx \in \alphabet^*} \sum_{\words' \in \alphabet^*} \phuman(\ctx \words')} \\
    & = 1.  
\end{align}
\end{subequations}
\end{proof}

\section{Linear Models}

\subsection{Rewriting Linear Models with Surprisal in Terms of PMI} \label{app:pmi-rewrite}

In this section, we demonstrate the equivalence between two linear models, one with surprisal and frequency as predictors, and the other with PMI and frequency as predictors. 
In particular, we consider a linear model of reading times $\func\left(\;\cdot \mid \pred(\ctx, \eosunit)\right)$, in which $\pred(\ctx, \eosunit) \colon \alphabet^* \times \eosalphabet \rightarrow \RD$ includes surprisal, frequency and additional baseline predictors $\predbase(\ctx, \eosunit) \colon \alphabet^* \times \eosalphabet \rightarrow \mathbb{R}^{D-3}$ and show that it is equivalent to a model in which surprisal is replaced with PMI. 
Consider the prediction $\rtpred(\ctx, \eosunit)$ which is the expected value of $\func\left(\;\cdot \mid \pred(\ctx, \eosunit)\right)$. To demonstrate the equivalence, we simply add and subtract an additional frequency term:
\begin{subequations}
\begin{align}
     \rtpred(\ctx, \eosunit) &= \bzero + \bfreq \unigramsurprisal(\eosunit) + \bsurp \surprisal(\ctx, \eosunit) + \paramsbase \predbase(\ctx, \eosunit)  \\
    & = \bzero + (\bfreq + \bsurp) \unigramsurprisal(\eosunit) - \bsurp (\underbrace{\unigramsurprisal(\eosunit) - \surprisal(\ctx, \eosunit)}_{=\pmi(\ctx, \eosunit)}) + \paramsbase \predbase(\ctx, \eosunit) \\
    & = \bzero + (\bfreq + \bsurp) \unigramsurprisal(\eosunit) -\bsurp \pmi(\ctx, \eosunit) + \paramsbase \predbase(\ctx, \eosunit), 
\end{align}
\end{subequations}
where $\params = [\bzero, \bfreq, \bsurp, \params_{\text{b}}] \in \RD$.

\subsection{The Orthogonalized Surprisal Model}\label{sec:orthogonalized-linear}

Under a linear model of reading time which includes orthogonal surprisal, frequency and (possibly) other predictors $\predbase(\ctx, \eosunit)$, reading time predictions are obtained as follows:
\begin{subequations}
\begin{align}
    \rtpred(\ctx, \eosunit)  & = \bzero + \bsurp \left(\surprisal(\ctx, \eosunit) - \frac{\cov(\surprisalrv, \frequencyrv)}{\cov(\frequencyrv, \frequencyrv)} \unigramsurprisal(\eosunit) \right) + \bfreq \unigramsurprisal(\eosunit) + \paramsbase \predbase(\ctx, \eosunit) \\
    & = \bzero + \bsurp \surprisal(\ctx, \eosunit)  + \left(\bfreq - \bsurp \frac{\cov(\surprisalrv, \frequencyrv)}{\cov(\frequencyrv, \frequencyrv)} \right) \unigramsurprisal(\eosunit) + \paramsbase \predbase(\ctx, \eosunit).
\end{align}
\end{subequations}
We estimate $\cov(\surprisalrv, \frequencyrv)$ (and $\cov(\frequencyrv, \frequencyrv)$) through the unbiased sample covariance on a training corpus $\corpus \defeq \{(\pred(\ctx_n, \eosunit_n), \rtn)\}_{n=1}^N$:
\begin{align}
    \frac{1}{N-1} \sum_{n=1}^N (\surprisal(\ctx_n, \eosunit_n) - \samplemean_\surprisal)(\unigramsurprisal(\eosunit_n) - \samplemean_\unigramsurprisal),
\end{align}
where $\samplemean_\surprisal$ and $\samplemean_\unigramsurprisal$ are the sample means computed over $\corpus$ for surprisal and frequency, respectively. In words, this means that we only use the word--context pairs present in the training data for approximating the inner products in \cref{eq:orthogonalized-predictor}. Recall that $\hilbertspace$ is infinite-dimensional, and we are not aware of efficient algorithms for exact computations of inner products on $\hilbertspace$.

\section{Further Technical Details on the Hilbert Space}
In this section we provide formal guarantees for the method presented in \cref{sec:disentangling}. In \cref{sec:hilbert-space-proof}, we justify why the Hilbert space introduced in \cref{sec:disentangling} is indeed a Hilbert space. In \cref{sec:projection}, we show that the projection in \cref{eq:orthogonalized-predictor} yields a predictor which is orthogonal to frequency.

\subsection{Predictor Variables as Elements of a Hilbert Space}\label{sec:hilbert-space-proof}

Let $(\alphabet^* \times \eosalphabet, \power(\alphabet^* \times \eosalphabet), \measure)$ be a probability space.
We further require that $\prefixhuman(\ctx) \phuman(\eosunit \mid \ctx) > 0$ for all $\ctx \in \alphabet^*$ and $\unit \in \alphabet$, which is equivalent to the assumption that $\phuman(\words) > 0$ for all $\words \in \alphabet^*$ by \cref{eq:autoregressive,eq:autoregressive-eos}.
We construct a Hilbert space $\hilbertspace$ over $\reals$ of all random variables of type $\genericrv \colon \alphabet^* \times \eosalphabet \rightarrow \reals$ with the restriction that $\expectation[\genericrv^2] < \infty$. We require the second moment to be finite since $\infty \notin \reals$.
In this paper, we are particularly interested in the random variables $\surprisalrv(\ctx, \eosunit) = \surprisal(\ctx, \eosunit)$, $\pmirv(\ctx, \eosunit) = \pmi(\ctx, \eosunit)$, $\frequencyrv(\ctx, \eosunit) = \unigramsurprisal(\eosunit)$, for $\ctx \in \alphabet^*$ and $\eosunit\in\eosalphabet$. They encode the distributions over surprisal, frequency and PMI, respectively, and are all distributed according to $\measure$. We have
\begin{subequations}
\begin{align}
\prob(\surprisalrv = \iota) = \sum_{\substack{\ctx \in \alphabet^*, \\ \eosunit \in \eosalphabet}}  \prefixhuman(\ctx) \phuman(\eosunit \mid \ctx) \mathbbm{1}\{  \iota = \surprisal(\ctx, \eosunit)\} \label{eq:prob-surprisal} \\
\prob(\frequencyrv = \upsilon)  = \sum_{\substack{\ctx \in \alphabet^*, \\ \eosunit \in \eosalphabet}} \prefixhuman(\ctx) \phuman(\eosunit \mid \ctx) \mathbbm{1}\{  \upsilon = \unigramsurprisal(\eosunit)\} \label{eq:prob-freq}\\
\prob(\pmirv = \mu) = \sum_{\substack{\ctx \in \alphabet^*, \\ \eosunit \in \eosalphabet}} \prefixhuman(\ctx) \phuman(\eosunit \mid \ctx) \mathbbm{1}\{  \mu = \pmi(\ctx, \eosunit)\}. \label{eq:prob-pmi} 
 \end{align}
 \end{subequations}
We define the following inner product:
\begin{subequations}
\begin{align}\label{eq:inner-product-derived}
    \langle \genericrv, \genericrvy \rangle & \defeq \expectation\left[\genericrv \genericrvy\right] \\
    & = \sum_{\substack{\ctx \in \alphabet^*, \\ \eosunit \in \eosalphabet}} \prefixhuman(\ctx) \phuman(\eosunit \mid \ctx) \genericrv(\ctx, \eosunit) \genericrvy(\ctx, \eosunit). 
\end{align}
\end{subequations}
Consequently, the norm is given by $\norm{\cdot} \defeq \sqrt{\langle \cdot, \cdot \rangle}$ and the distance metric between two random variables $\genericrv$ and $\genericrvy$ is $\distance(\genericrv, \genericrvy) \defeq \norm{\genericrv - \genericrvy}$.\footnote{Without the constraint that $\phuman(\words) > 0$ for all $\words \in \alphabet^*$, we would only obtain a seminorm.} 
With this choice of inner product, $\hilbertspace$ is a Hilbert space. That is because $\hilbertspace$ is the $L^2(\alphabet^* \times \eosalphabet)$ space, and
any such $L^2$ space is a Hilbert space \citep[p. 78]{rudin1987realcomplex}.

\subsection{Orthogonal Projection}\label{sec:projection}

In this section, we use the Hilbert projection theorem to show that frequency and orthogonalized surprisal are disentangled, i.e., that $\langle \frequencyrv,  \proj_{\frequencyrvcomp}(\surprisalrv) \rangle = 0$. We introduce the Hilbert projection theorem in \cref{prop:hilbert-projection} and apply it to our technique in \cref{thm:decorrelated}.

\begin{prop}[Hilbert Projection Theorem]\label{prop:hilbert-projection}
    Let $\hilbertspace$ be a Hilbert space and $\hilbertsubset \subseteq \hilbertspace$ be a nonempty closed convex set. For every $\genericrv \in \hilbertspace$ there exists a unique $\norm{\genericrv - \genericrvy'}$ which is equal to
    \begin{equation}
        \inf_{\genericrvy \in \hilbertsubset} \norm{\genericrv - \genericrvy}.
    \end{equation}
    If $\hilbertsubset$ is a closed vector subspace of $\hilbertspace$, then $\genericrv - \genericrvy'$ is orthogonal to $\hilbertsubset$, i.e., $\langle \genericrv - \genericrvy', \genericrvy \rangle =0$ for all $\genericrvy \in \hilbertsubset$.
\end{prop}
\begin{proof}
See \citet[pp. 306-9]{rudin1991functional}.
\end{proof}

\begin{prop}[Decorrelated Predictors]\label{thm:decorrelated}
    Let $\hilbertspace$ be the Hilbert space introduced in \cref{sec:disentangling} and \cref{sec:hilbert-space-proof}. Further, consider $\genericrv, \genericrvz \in \hilbertspace$ and define
    \begin{equation}
        \proj_{\genericrvz^\perp}(\genericrv)  \defeq \genericrv -  \frac{\langle \genericrv,  \genericrvz \rangle}{\langle \genericrvz, \genericrvz \rangle } \genericrvz.
    \end{equation}
    Then,
    \begin{equation}
        \langle \proj_{\genericrvz^\perp}(\genericrv), \genericrvz \rangle = 0.
    \end{equation}
\end{prop}

\begin{proof}
    Let $\hilbertsubset$ be the set of vectors spanned by $\genericrvz$, i.e., $\hilbertsubset \defeq \{ a\genericrvz \mid a \in \reals\}$. 
It is easy to see that $\hilbertsubset$ is a closed subspace of $\hilbertspace$. 
Then, by \cref{prop:hilbert-projection}, there exists a $\genericrvy'$ such that
    \begin{equation}
         \inf_{\genericrvy \in \hilbertsubset} \norm{\genericrv - \genericrvy} = \norm{\genericrv - \genericrvy'}.
    \end{equation}
By the definition of $\hilbertsubset$, we have $\genericrvy' = a' \genericrvz$ for some $a' \in \reals$. 
Because the infimum is achieved, to determine $a'$, we consider the following convex optimization problem:
    \begin{subequations}
    \begin{align}
\argmin_{a \in \reals}  \normsq{\genericrv - a \genericrvz} & =  \argmin_{a \in \reals} \normsq{a\genericrvz} - 2 \langle a\genericrvz, \genericrv \rangle + \normsq{\genericrv} \\
        & =  \argmin_{a \in \reals} a^2\normsq{\genericrvz} - 2a \langle \genericrvz, \genericrv \rangle. \label{eq:convex-in-scalar}
    \end{align} 
    \end{subequations}
Because \cref{eq:convex-in-scalar} is differentiable in $a$, we check the first-order optimality conditions
    \begin{equation}
        2a \normsq{\genericrvz} - 2 \langle \genericrvz, \genericrv \rangle = 0,
    \end{equation}
    so 
    \begin{subequations}
    \begin{align}
        \argmin_{a \in \reals} a^2\normsq{\genericrvz} - 2a \langle \genericrvz, \genericrv \rangle & = \frac{\langle \genericrvz, \genericrv \rangle}{\normsq{\genericrvz}} \\
        & = \frac{\langle \genericrvz, \genericrv \rangle}{\langle \genericrvz, \genericrvz \rangle}.
    \end{align}
    \end{subequations}
We note that this solution is unique because \cref{eq:convex-in-scalar} is convex in $a$.
    Thus, observing by \cref{prop:hilbert-projection} that
    \begin{equation}
        \langle \genericrv - \frac{\langle \genericrvz, \genericrv \rangle}{\langle \genericrvz, \genericrvz \rangle} \genericrvz, \genericrvz \rangle = 0,
    \end{equation}
we have the desired result that $\langle \proj_{\genericrvz^\perp}(\genericrv), \genericrvz \rangle = 0$.
\end{proof}

\section{Our Method in Relation to Residualization}\label{sec:residualization}

The technique presented in \cref{sec:disentangling} closely resembles another method used to decorrelate predictors---residualization (see, e.g., \citealp{kuperman2008,FLORIANJAEGER201023,Garcia2019-om}). Consider a linear regression setting with response $\responsevar\in\realsn$ and design matrix $\predmatrix \in \realsntwo$, i.e., 
$\nsamples$ data points for two predictors $\predone, \predtwo \in \realsn$, being column vectors. The idea behind residualization is to decorrelate the predictors by replacing one of them, say $\predone$, by the residuals obtained from the ordinary least squares solution of the regression model in which $\predone$ is the response and $\predtwo$ is the (only) predictor. The new predictor will thus take the value
\begin{align}\label{eq:residualization}
    \predresid = \predone - (\paramestzero + \paramestone \predtwo),
\end{align}
where $\paramestzero, \paramestone \in \reals$ are the ordinary least squares estimates for the intercept and slope, respectively, when regressing $\predone$ against $\predtwo$. Note that $\paramestzero=0$ and $\paramestone=\frac{\predone^{\intercal} \predtwo}{\predtwo^{\intercal}\predtwo}$ under mean centering.
In that case, \cref{eq:residualization} mirrors \cref{eq:orthogonalized-predictor}:
\begin{align}\label{eq:residualization-centered}
    \predresid = \predone - \frac{\predone^{\intercal} \predtwo}{\predtwo^{\intercal}\predtwo} \predtwo.
\end{align}
In \cref{eq:orthogonalized-predictor} we had the inner product between two vectors in the Hilbert space $\hilbertspace$, but here we have the inner product between two vectors in Euclidean space ($\predone$ and $\predtwo$), which is the dot product.
Indeed, residualization is defined only over a \emph{sample} of data points.
Our technique can thus be viewed as a functional generalization of residualization.
From a statistical perspective, \cref{eq:residualization-centered} provides estimates of residual values which can, ideally, be generalized to data beyond the sample of $\nsamples$ data points from which they were estimated. In a Hilbert space, on the other hand, there is no question of generalization. For the predictors we use, which are derived from a language model, we \emph{know} their true values. We only approximate the inner products in \cref{eq:orthogonalized-predictor} since computing their exact value would be intractable. That is, the reason for our approximation, which yields the same formula as residualization, is computational, rather than statistical.
 
\paragraph{Arguments Against Residualization.} Residualization has received criticism as a way to obtain more interpretable model coefficients \citep{york2012residualization, wurmfisicaro2014residualizing}. Consider the following three linear regression models estimated by ordinary least squares: 
\begin{align}
    \text{Model A}\colon\quad & \responsevar = \beta^A_0 + \beta^A_1 \predone + \beta^A_2\predtwo + \boldsymbol{\varepsilon} \\
    \text{Model B}\colon\quad & \responsevar = \beta^B_0 + \beta^B_1 \predresid + \beta^B_2\predtwo + \boldsymbol{\varepsilon} \\
    \text{Model C}\colon\quad & \responsevar = \beta^C_0 + \beta^C_2\predtwo + \boldsymbol{\varepsilon}, 
\end{align}
where $\boldsymbol{\varepsilon} \in \realsn$ is a vector of Gaussian noise variables. In the models above, we have that $\beta^A_1=\beta^B_1$ and $\beta^A_2 \neq \beta^B_2 = \beta^C_2$ \citep{wurmfisicaro2014residualizing}. That is, the effect of the residualized predictor---$\predone$ in the example above---remains the same after residualization (Model A vs. Model B). On the other hand, the estimated effect of the residualizing predictor on the response---$\predtwo$ in the example above---changes between Model B which regresses $\responsevar$ on $\predresid$ and $\predtwo$ and Model A which regresses $\responsevar$ on $\predone$ and $\predtwo$. The estimated effect of $\predtwo$ in Model B instead becomes equal to what it would have been under a model with a single predictor, regressing only on $\predtwo$ (Model C). These outcomes are contrary to what one might expect: The effect of the modified, residualized predictor $\predone$ is the same while the effect of the untouched predictor $\predtwo$ changes. This has indeed been a source of confusion in the literature \citep{wurmfisicaro2014residualizing}.
However, in our case, the fact that $\beta^A_2 \neq \beta^B_2 = \beta^C_2$ is actually the desired consequence: We sought to estimate the effect of context that is not correlated with frequency. In other words, we want the covariance between frequency and surprisal to be attributed to frequency, as it would be in a model that only regresses reading time on frequency (corresponding to Model C).
Moreover, we argue that the estimated coefficient is the wrong quantity to look at when measuring importance---it is the role of the predictor in relation to the others that should matter. While $\beta^A_1=\beta^B_1$ does not indicate a difference in importance, a measure of explained variance like LMG does, as we demonstrate by our results in \cref{fig:results-exp2}. We thus advocate for the use of such a measure instead of analyzing coefficients. 
Another remark relates to whether a residualized variable is interpretable as anything at all.
For instance, \citet{Breaugh2006} gives an example of height and weight of basketball players: Residualizing height with respect to weight would result in a residualized predictor that would involve a notion of height disentangled from weight, which is tricky to conceptualize in a real-world context.
However, in our case, we are addressing the question of what predictor should be extracted in the first place, i.e., from a language model as a stand-in for context. Our paper argued that orthogonalized surprisal gives a better interpretation for the effect of context than contextual surprisal does.

We hope that our work and this discussion can help shed further light on when residualization is, and is not, suitable.

\section{Dataset and Predictor Variables} \label{app:methods}

\subsection{Dataset} \label{app:dataset}

We use the Multilingual Eye-movement Corpus (MECO; \citealp{siegelman2022expanding}), which consists of eye-tracking-based reading-time data across 13 different languages for 12 Wikipedia-style articles about various topics. The articles have been carefully constructed to contain the same content across languages. Word-level reading time is recorded for between $32{-}54$ participants per language, using several different reading variables. Of these, for our main experiments, we use \defn{gaze duration}, which is the total fixation time on a word during its first pass (i.e., before the first time the gaze leaves the word). However, in \cref{fig:dll-languages} below we also show results for two other reading metrics: \defn{first fixation duration}, which is the duration of the first fixation that lands on the word, and \defn{total fixation duration}, which is the sum over all fixation durations of a word. In our experiments, we average the reading times for each word across all participants, excluding the participants that skipped (i.e., did not fixate on) the word. Thus, we avoid modeling the variance stemming from whether a word is skipped.

\subsection{Predictors} \label{app:regressions}

Our predictor variables are estimated in the following way: Surprisal estimates are obtained from mGPT \citep{mgpt2024}, which is a multilingual variant of GPT-3 that was trained on Wikipedia and the C4 corpus \citep{C4-corpus-2020}. It supports 61 languages, which intersected with the MECO dataset yields 11 languages for our experiments: Dutch, English, Finnish, German, Greek, Hebrew, Italian, Korean, Russian, Spanish, and Turkish. Estonian and Norwegian, which are present in MECO, are unfortunately not supported by mGPT. For each word in MECO, we sum the surprisals estimated by mGPT for each of the tokens that make up that word.
We use the estimates from \citet{robyn_speer_2022_7199437} to obtain word frequency (i.e., unigram surprisal) and length, following previous work. Finally, PMI estimates are obtained from surprisal and frequency through \cref{eq:pmi-decomposition}. All predictors are standardized (i.e., set to mean zero and standard deviation one) before computing orthogonalized surprisal values and fitting the regression models.

\section{Additional Experiments}\label{sec:additional-experiments}
We complement the main experiments presented in \cref{sec:experiments} with three additional empirical analyses. In \cref{sec:orthogonal-frequency}, we take an alternative position to the one in the main text, using an orthogonalized \emph{frequency} predictor.
In \cref{app:no-length}, we exclude the length predictor and perform the same analysis as we did in the main text. In \cref{sec:predictive-power-experiments}, we compare the predictive power of  nonlinear models across different sets of predictors, including analyses on first fixation duration and total fixation duration.

\subsection{Orthogonalized Frequency}\label{sec:orthogonal-frequency}

\begin{figure*}
    \centering
    \includegraphics[width=\textwidth]{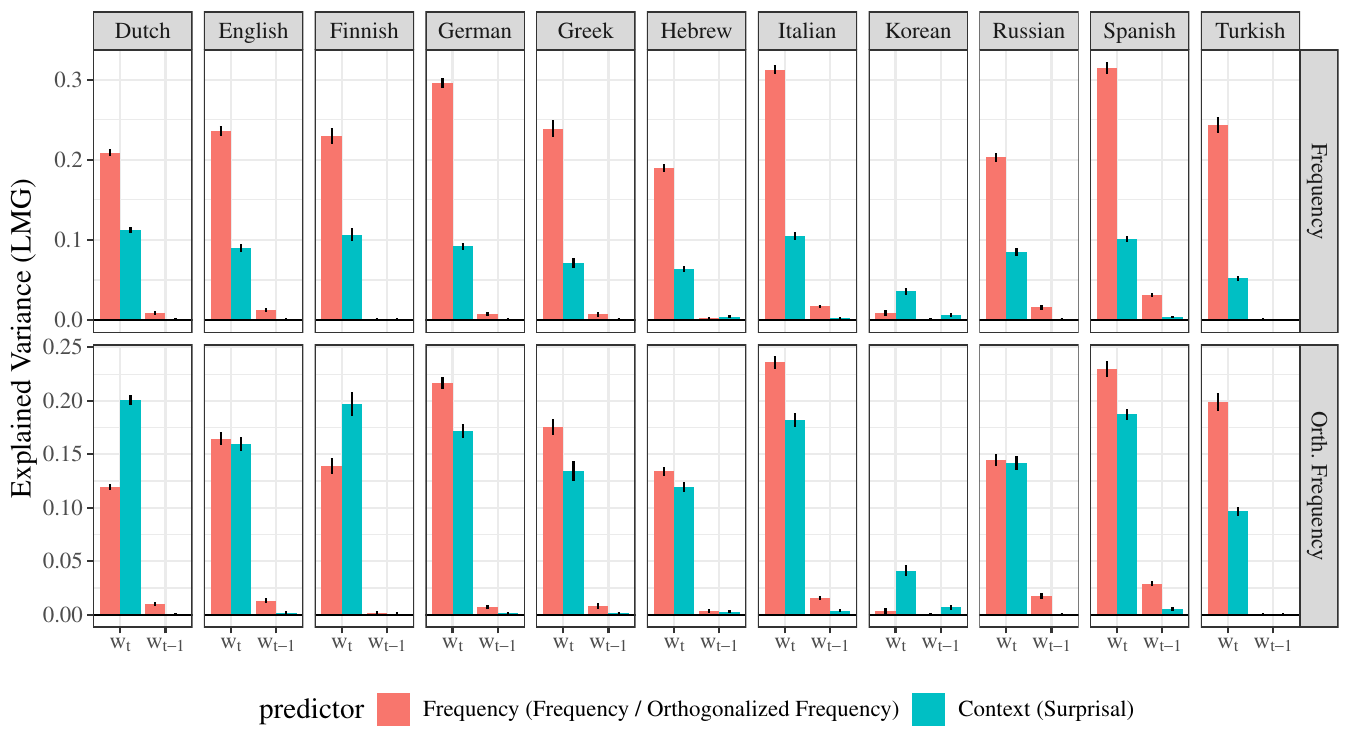}
    \caption{
    This figure is analogous to \cref{fig:results-exp2}, with the difference that frequency is projected onto the orthogonal complement of surprisal, resulting in an orthogonalized frequency predictor. Even when the shared effect of frequency and surprisal is attributed to surprisal, we still find that the frequency predictor explains more variance than surprisal for most languages.}
    \label{fig:orthogonalized-freq}
\end{figure*}

Here, we provide an additional analysis in which we derive a new frequency predictor, swapping the two variables in \cref{eq:orthogonalized-predictor}. In this case, the shared effect of frequency and surprisal on reading time is attributed to surprisal, and the frequency effect represents the effect beyond what is already explained by surprisal. We present the results in \cref{fig:orthogonalized-freq}.
Comparing this analysis to \cref{fig:results-exp2} may be considered analogous to \citeposs{shain-2019-large} study, which compares the independent effects of surprisal and frequency by adding them in as predictors on top of baseline models that contain the other. However, in contrast to \citet{shain-2019-large}, we find that frequency appears to be more important than contextual effects in explaining reading times; for most languages, frequency remains the more important predictor even after orthogonalization.

\subsection{Results without Length}\label{app:no-length}
\begin{figure*}
    \centering
    \vspace{-6pt}
    \includegraphics[width=\textwidth]{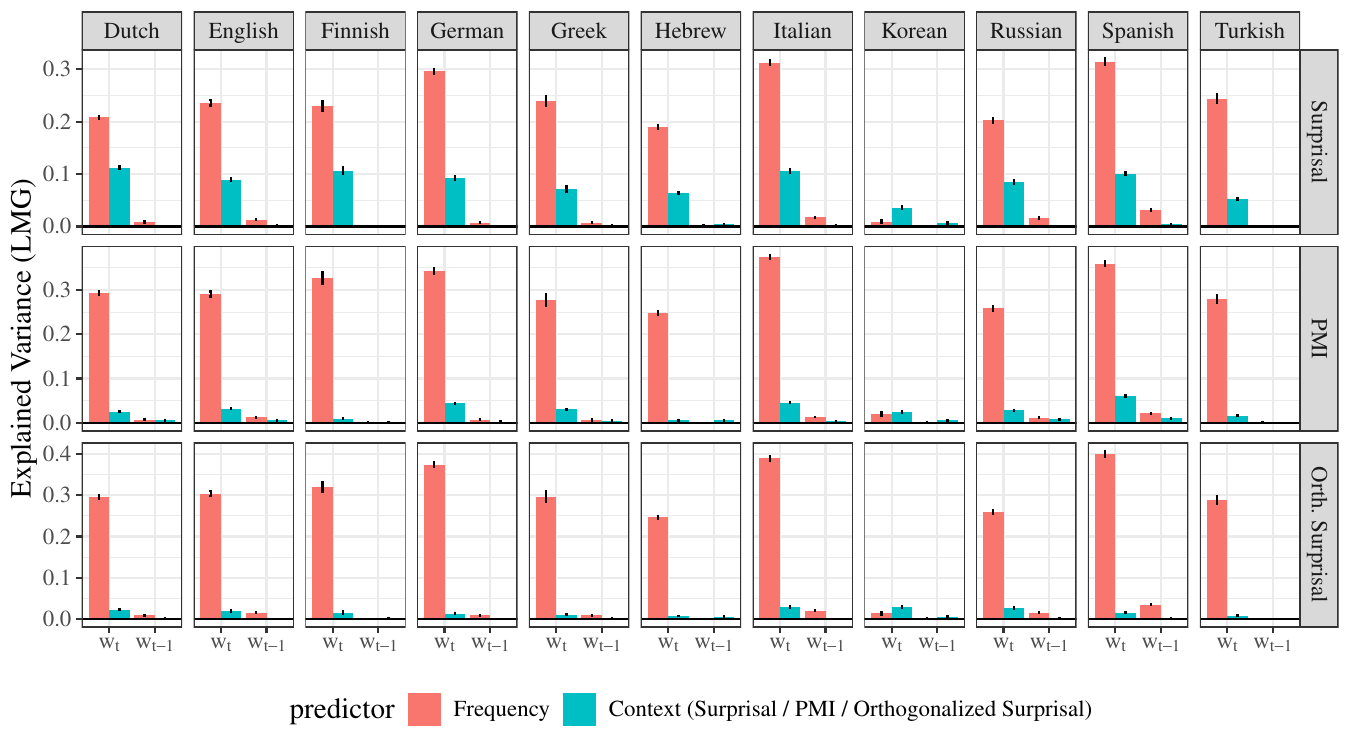}
    \vspace{-24pt}
    \caption{
    This figure is analogous to \cref{fig:results-exp2}, except that it shows results when excluding the word length predictor.
    }
    \vspace{-12pt}
    \label{fig:no-length}
\end{figure*}

We complement the experiments in \cref{sec:experiments} with an additional analysis which excludes length. We follow the same experimental setup as described in the main portion of the text, this time with three linear models that include predictors: (i) surprisal and frequency, (ii) PMI and frequency, and (iii) orthogonalized surprisal and frequency. We include spillover variables from the previous word. The results are shown in \cref{fig:no-length}. 
We observe that the implications on context found in \cref{fig:results-exp2} do not change when excluding length. 

\subsection{Psychometric Predictive Power}\label{sec:predictive-power-experiments}
\begin{figure}[t]
    \centering
    \vspace{-6pt}
    \includegraphics[width=\columnwidth]{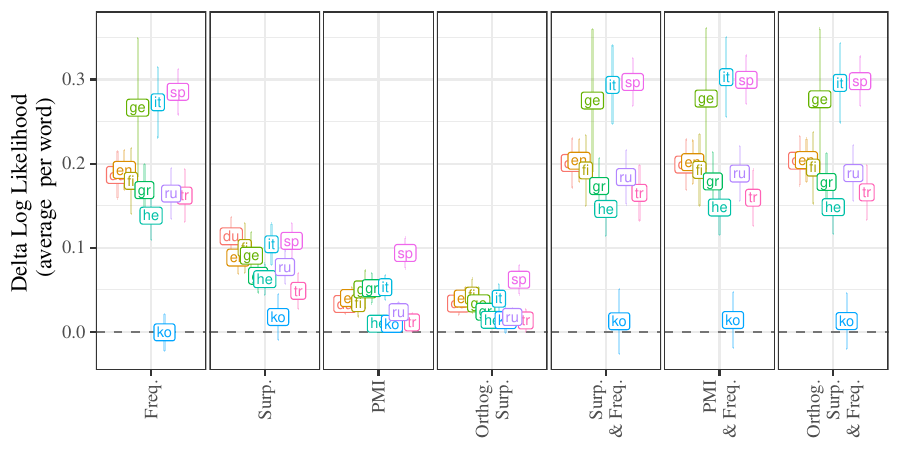}
    \vspace{-16pt}
    \caption{\dll of GAM models. Models include only the indicated predictors. Error bars are 95\% confidence intervals across folds of data.
    }
    \vspace{-12pt}
    \label{fig:results-exp1}
\end{figure}

While the surprisal and PMI models are equivalent under a linear model, that relationship does not necessarily need to hold under a nonlinear one.
Therefore, we compare the psychometric predictive power by fitting generalized additive models (GAMs). GAMs are a class of additive models that can learn non-linear relationships between predictor and response variables. All the terms in our GAMs are spline-based smooth terms, that include either a contextual predictor variable (i.e., surprisal, PMI, or orthogonalized surprisal), or frequency. We restrict our smooth terms to six basis functions, following the logic outlined by \citet{hoover-etal-2023-plausibility}. GAMs are fit using the \texttt{mgcv} package in \texttt{R}. Two example calls are given below:

\lstset{frame=none,
  language=R,
  showstringspaces=false,
  columns=flexible,
  numbers=none,
  keywords={lmer, data, te, bs, s, gam},
  basicstyle={\small\ttfamily},
  keywordstyle=\color{Bittersweet},
  stringstyle=\color{Bittersweet},
  commentstyle=\color{Bittersweet},
  breaklines=true,
  breakatwhitespace=true,
  tabsize=3}

\newcommand{\listlingr}[1]{{\small\ttfamily#1}}

\begin{lstlisting}
gam(reading_time ~ s(surprisal, bs = 'cr', k = 6) + s(prev_surprisal, bs = 'cr', k = 6), data = .)

gam(reading_time ~ s(pmi, bs = 'cr', k = 6) + s(prev_pmi, bs = 'cr', k = 6) + s(frequency, bs = 'cr', k = 6) + s(prev_frequency, bs = 'cr', k = 6), data = .)

\end{lstlisting}

\noindent We consider an additional baseline model which is the average reading time estimated from the training set. 
We compare delta log-likelihood \dll---the average difference in likelihood between the target models mentioned above and the baseline model---as estimated over ten folds of cross-validation across several different sets of predictor variables. 

Results for gaze duration are visualized in \cref{fig:results-exp1}. We observe three big trends: First, we find that all predictors lead to positive \dll, except for the case of Korean, indicating that they are useful for predicting reading times. However, second, when looking at models with just one predictor variable, we observe that frequency alone leads to higher \dll than surprisal and PMI do, and that surprisal and PMI tend to result in higher \dll than orthogonalized surprisal. This is to be expected. We know from prior research that frequency plays a large role in explaining by-word processing effort, and because orthogonalized variables, by definition, are decorrelated with frequency, they are not expected to be strong predictors of reading times, alone.

The right three facets of \cref{fig:results-exp1} show models that combine contextual and non-contextual predictors into a single model. Here, we observe only insignificant, nearly invisible differences between the models' $\dll$.
We conclude that all three implementations of context are equally good at predicting reading times.

\begin{figure*}[h]
    \vspace{-10pt}
    \centering
    \includegraphics[width=\textwidth]{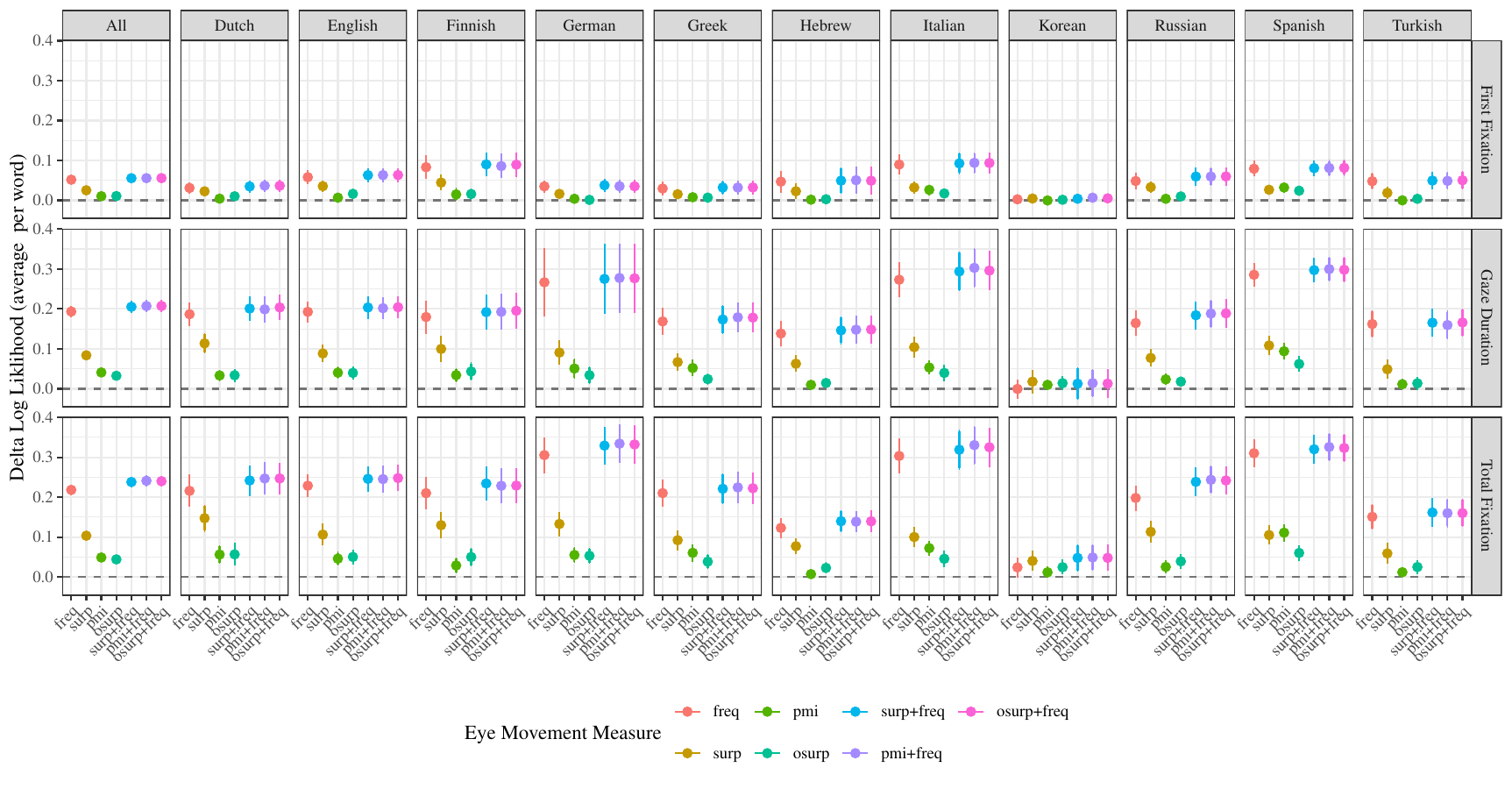}
    \caption{\dll of GAM models over three different aggregated measurements of reading time. Error bars are 95\% confidence intervals across folds of data.}
    \label{fig:dll-languages}
\end{figure*}

In \Cref{fig:dll-languages}, we show our generalized additive modeling results, broken down by language, across the $x$-facets. We also show results for reading time metrics other than gaze duration, including first fixation duration (top row) and total reading times (bottom row). These results are consistent with those reported in \cref{fig:results-exp1}. We find that of the four individual predictors, frequency leads to the highest \dll, followed by surprisal, PMI, and then orthogonalized variants. When combining our non-contextual predictor (frequency), alongside these contextual predictors, we do not observe differences in \dll. 

\section{Connection with Model Size}\label{sec:discussion}

The results presented in this article may help explain a trend recently observed in the computational psycholinguistics literature: Surprisal values of larger LMs provide a worse fit to human reading-time data compared to those of medium-sized models \citep{oh-schuler-2023-transformer,oh2022does}.
Specifically, \citet{oh-etal-2024-frequency} suggest that this is because larger models are incredibly accurate at predicting rare words in context. 
Medium-sized models, on the other hand, are not as good at predicting rare words in context. Therefore, surprisal estimates for these words are closer to their unigram frequencies, i.e., non-contextual surprisal. 
If reading times are primarily driven by frequency effects, as suggested by our analysis, the surprisal predictor should---on its own---yield stronger predictive power if it is closer to frequency, as is the case for medium-sized models. Thus, this could explain why the decoupling of surprisal and frequency in these larger models results in poorer fits to human reading times. 

Similarly, this could be a reason for why surprisal estimates derived from lossy contexts have been shown to be more predictive of reading times \citep{futrell2020lossy,kuribayashi-etal-2022-context}: Restricting the context might make the surprisal estimates more similar to unigram frequencies.

\end{document}